\documentclass[journal]{IEEEtran}
\usepackage{amsfonts,amssymb}
\usepackage{amsmath}
\usepackage{graphicx}
\usepackage{booktabs}
\usepackage{multirow}
\usepackage{color}
\usepackage{cite}
\usepackage{upgreek}
\usepackage[colorlinks=true,
            linkcolor=blue,
            anchorcolor=blue,
            citecolor=blue]{hyperref}
\usepackage{algorithmicx,algorithm}
\usepackage{graphicx}
\usepackage{algpseudocode}
\usepackage{epstopdf}
\usepackage{xcolor}
\usepackage{xpatch}

\ifCLASSINFOpdf
\else
\fi

\makeatletter
\def\changeBibColor#1{%
	\in@{#1}{VAE1}
	\ifin@\color{red}\else\normalcolor\fi
}

\xpatchcmd\@bibitem
{\item}
{\changeBibColor{#1}\item}
{}{}
\makeatother

\begin{document}
\title{Diffusion Models Meet Remote Sensing: Principles, Methods, and Perspectives}

\author{Yidan~Liu,
        Jun~Yue,~\IEEEmembership{Member,~IEEE},
        Shaobo~Xia,
        Pedram~Ghamisi,~\IEEEmembership{Senior Member,~IEEE},
        Weiying~Xie,~\IEEEmembership{Senior Member,~IEEE},
        and~Leyuan~Fang,~\IEEEmembership{Senior Member,~IEEE}
\thanks{This work was supported in part by the National Natural Science Foundation of China under Grant U22B2014, Grant 62101072 and Grant 62425109. (\textit{Yidan Liu and Jun Yue contributed equally to this work.}) (\textit{Corresponding author: Leyuan Fang.})}
\thanks{Yidan Liu and Leyuan Fang are with the College of Electrical and Information Engineering, Hunan University, Changsha 410082, China (e-mail: yidanliu@hnu.edu.cn; fangleyuan@gmail.com). \par
Jun Yue is with the School of Automation, Central South University, Changsha 410083, China (e-mail: junyue@csu.edu.cn). \par
Shaobo Xia is with the Department of Geomatics Engineering, Changsha University of Science and Technology, Changsha 410114, China (e-mail: shaobo.xia@csust.edu.cn). \par
Pedram Ghamisi is with the Helmholtz-Zentrum Dresden-Rossendorf (HZDR), Helmholtz Institute Freiberg for Resource Technology, 09599 Freiberg, Germany, and also with Lancaster University, LA1 4YR Lancaster, U.K. (e-mail: p.ghamisi@gmail.com). \par
Weiying Xie is with the State Key Laboratory of Integrated Services Networks, Xidian University, Xi'an 710071, China (e-mail: wyxie@xidian.edu.cn).}}

\markboth{IEEE TRANSACTIONS ON GEOSCIENCE AND REMOTE SENSING,~Vol.~X, No.~X, X~2024}{}%

\maketitle
\begin{abstract}
As a newly emerging advance in deep generative models, diffusion models have achieved state-of-the-art results in many fields, including computer vision, natural language processing, and molecule design. The remote sensing community has also noticed the powerful ability of diffusion models and quickly applied them to a variety of tasks for image processing. Given the rapid increase in research on diffusion models in the field of remote sensing, it is necessary to conduct a comprehensive review of existing diffusion model-based remote sensing papers, to help researchers recognize the potential of diffusion models and provide some directions for further exploration. Specifically, this paper first introduces the theoretical background of diffusion models, and then systematically reviews the applications of diffusion models in remote sensing, including image generation, enhancement, and interpretation. Finally, the limitations of existing remote sensing diffusion models and worthy research directions for further exploration are discussed and summarized.
\end{abstract}



%

\IEEEpeerreviewmaketitle

\section{Introduction}
%
%
%
%

\IEEEPARstart{R}{emote} sensing (RS), as an advanced earth observation technology, has been widely used in civilian and military fields such as environmental monitoring, urban planning, disaster response, and camouflage detection \cite{RS1,RS2,RS3,RS4,RS5}. Following the boom of artificial intelligence, employing deep learning models to interpret RS images has become a large-scale solution for these applications \cite{DLinRS}. Early intelligent RS interpretation methods primarily relied on supervised deep neural networks, which were trained with massive data and high-quality annotations. However, the scarcity of annotations and the high acquisition costs of RS images have hindered further advancements in these methods.

\begin{figure}[t]
\centering
\includegraphics[scale=0.46]{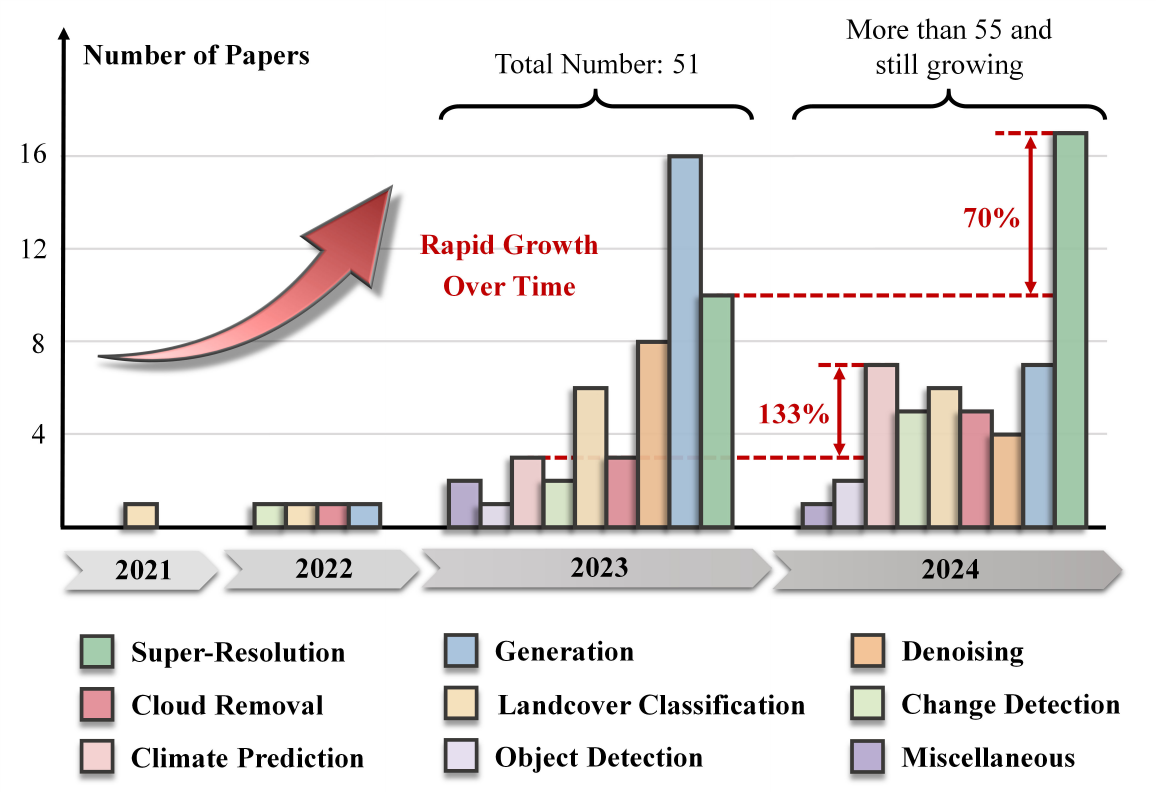}
\caption{\textcolor{black}{The number of papers on diffusion models for various RS tasks from 2021 to 2024. The data for 2024 is up to the second quarter. From the chart, it is evident that diffusion models firstly dominated the field of RS image generation and quickly expanded to more complex applications. For instance, the number of papers on climate prediction increased from 3 in 2023 to 7 in 2024, more than doubling. Similarly, the diffusion models of RS image super-resolution rose from 10 papers to 17 papers, an increase of 70\%.}}
\label{development}
\end{figure}

The advent of deep generative models effectively addresses the limitations of traditional supervised interpretation methods, bringing new opportunities for intelligent processing of RS images. These generative models can not only generate new RS images from limited data samples, but also generate annotation information directly from low-quality or unlabeled RS images by learning the mapping relationships between images, thereby reducing the reliance on high-quality manual annotations. Additionally, they also better in learning representations for image details and complex scenes than regular deep learning models. \textcolor{black}{For example, Variational Autoencoders (VAEs) are good at generating diverse RS images and learning meaningful representations \cite{VAE1, VAE2}. Generative Adversarial Networks (GANs) can produce highly realistic and visually appealing RS image samples \cite{GAN1, GAN2}. Normalizing Flows (NFs) can generate high-quality images and perform efficient inference \cite{NF1, NF2}. Despite the flourishing development of these generative models in the RS community, each comes with its limitations. VAE \cite{VAE} requires a trade-off between reconstruction loss (similarity between the output and the input) and latent loss (proximity of the hidden nodes to the normal distribution), so that the generated images are often blurry. The generated results of GAN \cite{GAN} are greatly influenced by the hyperparameters, and its training process is unstable, easy to fall into the pattern collapse (i.e., the generated sample pattern is singular and cannot cover diverse patterns). The structure of NF-based models \cite{NF} needs to comply with the calculation of probability density, resulting in limited scalability and flexibility.}


\begin{figure*}[t]
\centering
\includegraphics[scale=0.84]{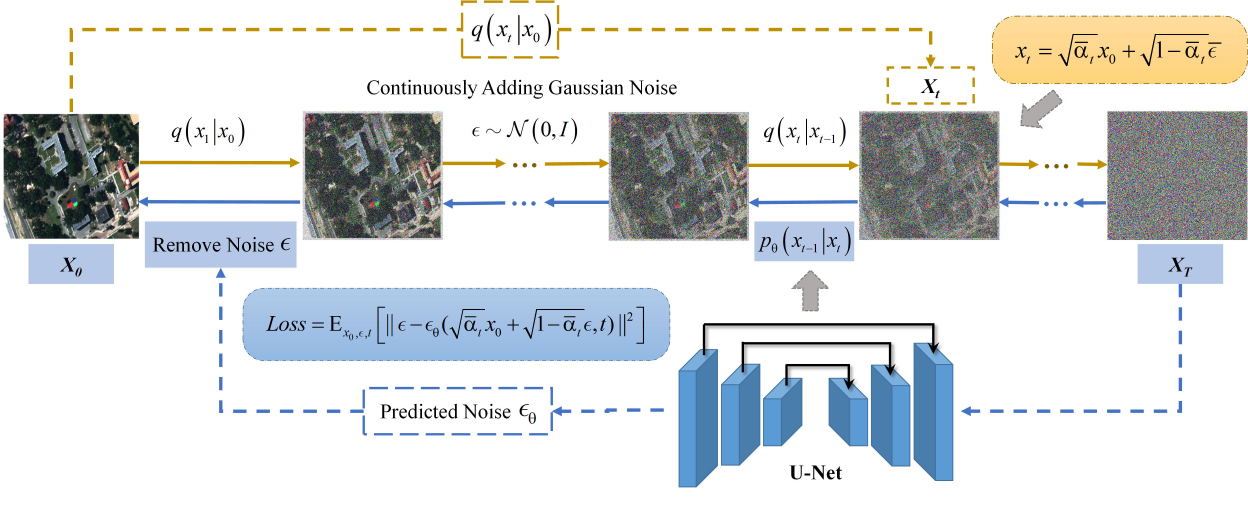}
\caption{The training procedure of denoising diffusion probabilistic model (DDPM), where yellow lines represent the forward diffusion process, and blue lines represent the backward diffusion process. Specifically, the network is used to fit the distribution $p(x_{t-1} | x_t)$ and output the predicted noise $\epsilon_{\theta}$. Then, minimize the distance between the predicted noise $\epsilon_{\theta}$ and the actual noise $\epsilon$ thereby optimizing the network and completing the training of diffusion model.}
\label{diffusion_process}
\end{figure*}

In this context, diffusion models \cite{DDPM}, as a newly emerging type of deep generative models, have brought about a revolutionary advancement in artificial intelligence. By modeling the inverse process of transforming regular images into random noise, diffusion models have demonstrated unprecedented performance in 2D and 3D image generation \cite{text-to-image0, DM3, 3dimage6, 3dimage4, 3dcvpr2024}, image editing \cite{editing, DM22, DM23, DM1}, image translation \cite{trans0, trans1, trans2, trans3}, and other computer vision tasks \cite{DM00,DM0,DM02,DM01}. Moreover, they have achieved state-of-the-art results in many other fields, including natural language processing \cite{NLP0,NLP1,NLP2}, audio synthesis \cite{AU0,AU1,AU2}, and molecular design \cite{molecular1,molecular2,molecular0}, challenging the long-standing dominance of GANs.

Given these remarkable achievements, the RS community has also quickly applied diffusion models to a variety of tasks for image processing. Since 2021, the application of diffusion models in RS has shown a rapid development trend of expanding scope and increasing quantity (see Fig. \ref{development}). In fact, diffusion models have significant advantages over other deep generative models in processing and analyzing RS images.

\begin{itemize}
\setlength{\parskip}{3pt}
    \item Firstly, due to the atmospheric interference and limitations of imaging equipment, RS images often contain noise. The inherent denoising ability of diffusion models can rightly eliminate these negative effects.\par
    
    \item Secondly, RS images are highly diverse due to differences in collection time, equipment, and environment. The architecture of diffusion models is flexible, allowing the introduction of conditional constraints to cope with various changes.

    \item Thirdly, RS images always contain diverse and complex scenes. The precise mathematical derivation and progressive learning process of diffusion models offer advantages in learning such complex data distribution. 

   \item Additionally, diffusion models can provide more stable training than GANs, which is suitable for training large-scale RS datasets.
\end{itemize}

In summary, diffusion models possess great development potential in the field of RS. Therefore, it is necessary to review and summarize existing diffusion model-based RS papers to help researchers gain a comprehensive understanding of the current research status, and identify the gaps in the application of diffusion models in RS, thereby promoting further development in this field.

The remainder of this paper is organized as follows. \textbf{Section \uppercase\expandafter{\romannumeral2}} introduces the theoretical background of diffusion models. \textbf{Section \uppercase\expandafter{\romannumeral3}} reviews the application of diffusion models across various RS image processing tasks, and demonstrates the superiority of diffusion models through a series of visual experimental results and quantitative metrics. \textbf{Section \uppercase\expandafter{\romannumeral4}} discusses the limitations of the existing RS diffusion models and reveals possible research directions in the future. Finally, conclusions are drawn in \textbf{Section \uppercase\expandafter{\romannumeral5}}.

\section{Theoretical Background of Diffusion Models}
Diffusion models, also known as diffusion probabilistic models, constitute a family of deep generative models. In essence, a generative model aims to transform a random distribution (i.e., noise) into a `probability distribution' that matches the distribution of the observed dataset, thereby producing desired outcomes through sampling from this `probability distribution'.

Obtaining the target distribution directly is evidently challenging. However, disrupting a regular distribution into random noise is comparatively straightforward and can be accomplished by continually adding Gaussian noise, as depicted in Fig. \ref{diffusion_process}. The concept of diffusion models is inspired by this process and involves learning the reverse denoising process, which dates back to 2015 \cite{DPM2015} and gained popularity following the publication of the denoising diffusion probabilistic model (DDPM) in 2020 \cite{DDPM}.

\subsection{Denoising Diffusion Probabilistic Model (DDPM)}
As shown in Fig. \ref{diffusion_process}, the training procedure involves two phases: the forward diffusion process and the backward diffusion process.

\textbf{Forward Diffusion Process:} Given the original image \( x_0 \), this process generates the noise-contaminated images \( x_1, x_2, \ldots, x_T \) through \( T \) iterations of noise addition. The image \( x_t \) obtained at each step is only related to \( x_{t-1} \). Thus, this process can be represented by a Markov chain:
\begin{gather}
q(x_t | x_{t-1}) = \mathcal{N}(x_t; \sqrt{1 - \beta_t}x_{t-1}, \beta_t I), \label{forward1}\\[2mm]
q(x_{1:T} | x_0) = \prod_{t=1}^{T}q(x_t | x_{t-1}) = \prod_{t=1}^{T} \mathcal{N}(x_t; \sqrt{1 - \beta_{t-1}}, \beta_t I),
\label{forward2}
\end{gather}
where $q(x_t | x_{t-1})$ is the transition probability of the Markov chain, which represents the distribution of Gaussian noise added at each step. $\beta_t$ is a hyperparameter for the variance of the Gaussian distribution, linearly increasing with $t$. $I$ denotes the identity matrix with the same dimensions as the input image \( x_0 \).

An important property of the forward process is that it allows to directly obtain any noised image \( x_t \) from the original image \( x_0 \) and $\beta_t$, which is achieved through the reparameterization technique \cite{VAE}. Specifically, with the notation of \(\alpha_t = 1 - \beta_t\) and \(\bar{\alpha}_t = \prod_{i=1}^{t} \alpha_i\), Eq. (\ref{forward1}) could be expanded as
\begin{align}
x_t &= \sqrt{\alpha_t}x_{t-1} + \sqrt{1 - \alpha_t}\epsilon_{t-1} \nonumber\\
&= \sqrt{\alpha_t} \left( \sqrt{\alpha_{t-1}}x_{t-2} + \sqrt{1 - \alpha_{t-1}}\epsilon_{t-2} \right) + \sqrt{1 - \alpha}_t\epsilon_{t-1} \nonumber\\
&= \sqrt{\alpha_t\alpha_{t-1}}x_{t-2} + \sqrt{1 - \alpha_{t-1} \alpha_{t-2}} \bar{\epsilon}_{t-2} \nonumber\\
&= \ldots \nonumber\\
&= \sqrt{\bar{\alpha}_t}x_0 + \sqrt{1 - \bar{\alpha}_t}\bar{\epsilon},
\label{forward3}
\end{align}
where $\epsilon_t$, $\epsilon_{t-1} \sim \mathcal{N}(0, I)$ and $\bar{\epsilon}_{t-2}$ is their merged result. According to the additivity of the independent Gaussian distribution, i.e., \(\mathcal{N}(0, \sigma_1^2 I) + \mathcal{N}(0, \sigma_2^2 I) \sim \mathcal{N}(0, (\sigma_1^2 + \sigma_2^2)I)\), the third line of Eq. (\ref{forward3}) conforms to Gaussian distribution, which means that the final derivation result also conforms to Gaussian distribution. Therefore, any noised image \( x_t \) satisfies:
\begin{equation}
q(x_t | x_0 ) = \mathcal{N}(x_t; \sqrt{\bar{\alpha}_t}x_0, 1 - \bar{\alpha}_t I).
\label{forward4}
\end{equation}
In this way, when $T \to \infty$, \( x_T \) can converge to the standard normal distribution $\mathcal{N}(0, I)$, consistent with the original design intention.

\textbf{Backward Diffusion Process:} This process aims to obtain the reversed transition probability $q(x_{t-1} | x_t)$, thereby gradually restoring the image \(\hat{x_0}\) from the noise. However, $q(x_{t-1} | x_t)$ is difficult to solve explicitly, so a neural network is employed to learn this distribution:

\begin{equation}
p_{\theta}(x_{t-1} | x_t) = \mathcal{N}(x_{t-1}; \mu_{\theta}(x_t, t), \Sigma_{\theta}(x_t, t)),
\label{backward1}
\end{equation}
where $\theta$ represents the parameters of the neural network to be optimized, and the network is typically based on a U-Net architecture \cite{unet}. Accordingly, the backward diffusion process can be expressed as
\begin{align}
p_{\theta}(x_{0:T}) &= p(x_T) \prod_{t=1}^{T} p_{\theta}(x_{t-1} | x_t).
\label{backward2}
\end{align}

The training goal of the network is to match the backward diffusion process \(p_{\theta}(x_0, x_1, \ldots, x_T)\) with the forward diffusion process \(p_{\theta}(x_0, x_1, \ldots, x_T)\), which can be achieved by minimizing the Kullback-Leibler (KL) divergence:
\begin{align}
{{\cal L}}(\theta ) &= \text{KL}(q(x_0, x_1, \ldots, x_T) || p_{\theta}(x_0, x_1, \ldots, x_T)) \nonumber\\[1mm]
&=-\mathbb{E}_{q(x_{0:T})}[\log p_{\theta}(x_0, x_1, \ldots, x_T)] + \text{const} \label{backward3} \\
&=-\mathbb{E}_{q(x_{0:T})}[-\log p(x_T) - \sum_{t=1}^T \log \frac{p_{\theta}(x_{t-1} | x_t)}{q(x_t | x_{t-1})}] + \text{const}. \nonumber
\end{align}
Here, $const$ denotes a constant independent of $\theta$, and the first term of Eq. (\ref{backward3}) represents the variational lower-bound of the negative log-likelihood, similar to VAE.

Notably, when the prior \( x_0 \) is introduced in $q(x_{t-1} | x_t)$, it can be converted by Bayes Rule:
\begin{align}
q(x_{t-1} | x_t, x_0) &= \frac{q(x_t, x_0, x_{t-1})}{q(x_t, x_0)} \nonumber \\[1mm]
&= \frac{q(x_0)q(x_{t-1} | x_0)q(x_t | x_{t-1}, x_0)}{q(x_0)q(x_t | x_0)} \nonumber \\[1mm]
&= q(x_{t} | x_{t-1}, x_0) \frac{q(x_{t-1} | x_0)}{q(x_t | x_0)},
\label{change1}
\end{align}
where $q(x_{t} | x_{t-1}, x_0)$ is defined in Eq. (\ref{forward1}), $q(x_{t-1} | x_0)$ and $q(x_t | x_0)$ can be obtained by Eq. (\ref{forward4}). After simplification, Eq. (\ref{change1}) can be rewritten as
\begin{gather}
q(x_{t-1} | x_t, x_0) = \mathcal{N}(x_{t-1}; \tilde{\mu}_t(x_t), \tilde{\beta}_t \mathbf{I}) \label{change2},\\[2mm]
\tilde{\mu}_t(x_t)  = \frac{1}{\sqrt{\alpha_t}} \left( x_t - \frac{1 - \alpha_t}{\sqrt{1 - \alpha_t}} \bar{\epsilon} \right), \label{change21}\\[2mm]
\hat{\beta}_t = \frac{1 - \alpha_{t-1}}{1 - \alpha_t} \beta_t.
\label{change22}
\end{gather}
$\alpha_t$ and $\beta_t$ are both constants in the above equation, only $\bar{\epsilon}$ in Eq. (\ref{change21}) can be parameterized by the neural network as
\begin{gather}
\mu_{\theta}(x_t, t)  = \frac{1}{\sqrt{\alpha_t}} \left( x_t - \frac{1 - \alpha_t}{\sqrt{1 - \bar{\alpha}_t}} \epsilon_{\theta}(x_t, t) \right).
\label{change23}
\end{gather}

In other words, the constructed neural network is to learn the noise $\epsilon_{\theta}(x_t, t)$, which is reasonable since the process from \( x_t \) to \( x_{t-1} \) is essentially a denoising process.

According to \cite{DDPM}, the optimization goal of the network can be further simplified with the help of Eq. (\ref{change23}) and (\ref{forward3}) as the following form:
\begin{gather}
{{\cal L}}_{simple}(\theta ) = \mathbb{E}_{x_0,\epsilon, t} \left[ \left\| \epsilon - \epsilon_{\theta} \left( \sqrt{\bar{\alpha}_t}x_0 + \sqrt{1 - \bar{\alpha}_t} \epsilon, t \right) \right\|^2 \right],
\label{loss_simp}
\end{gather}
which intuitively shows that the core of the diffusion model is to minimize the distance between the predicted noise $\epsilon_{\theta}$ and the actual noise $\epsilon$.

\begin{figure}[t]
\centering
\includegraphics[scale=0.4]{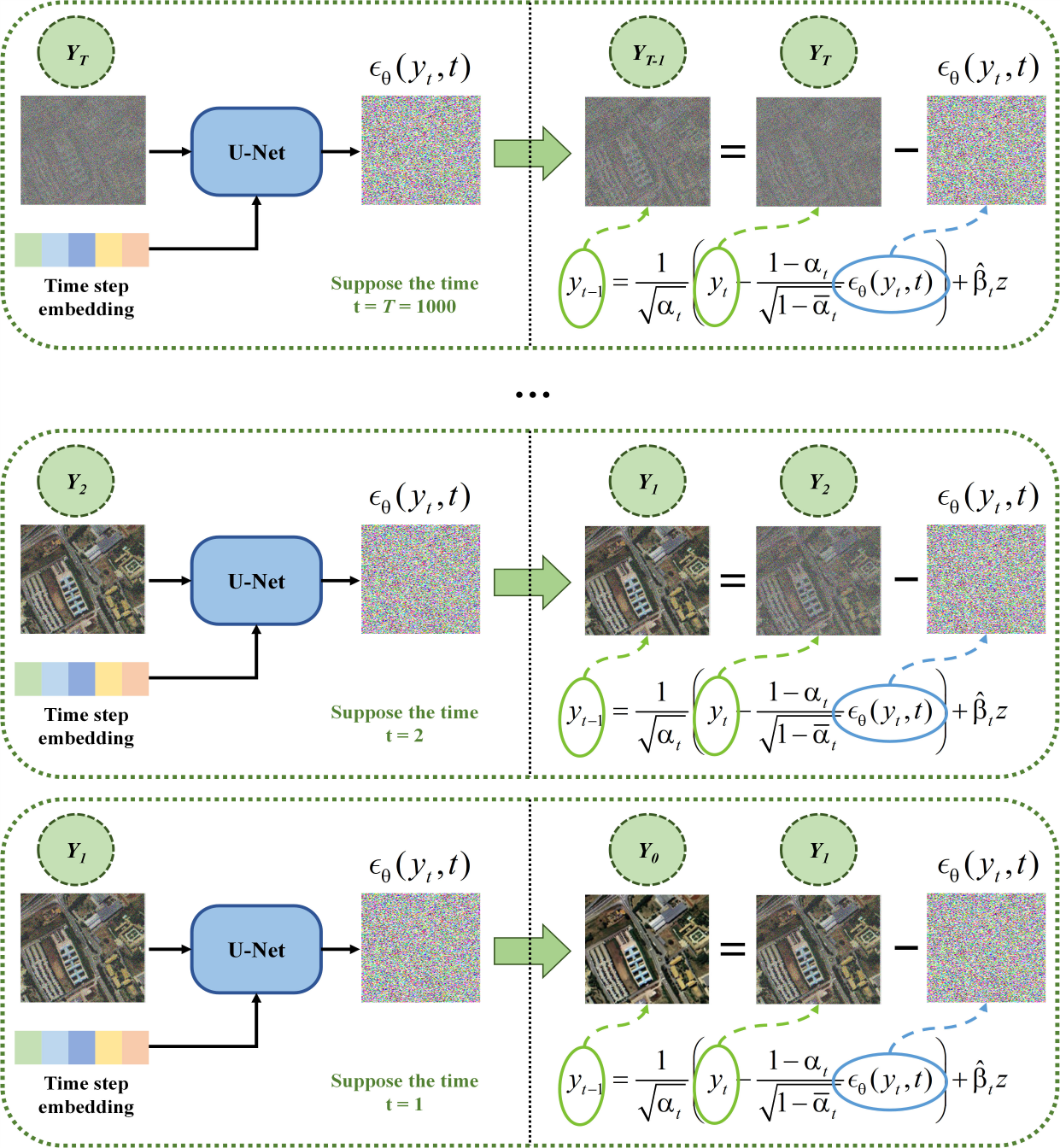}
\caption{The sampling process of DDPM. Supposing that sampling begins at T=1000, the noise distribution $\epsilon_{\theta}(y_t, t)$ is obtained from the well-trained diffusion model. Then, the noised image $Y_t$ is used to subtract the noise $\epsilon_{\theta}(y_t, t)$, resulting in a denoised image $Y_{t-1}$. This denoised image $Y_{t-1}$ is then input into the diffusion model to obtain the noise image for the next timestep. This process is repeated until \(t = 1\), at which point the denoised image is quite clear.}
\label{sampling_process}
\end{figure}

\textbf{Sampling Process:} In the inference process, which is also known as the sampling process, a new image \(y_0\) can be generated from either Gaussian noise or a noisy image \(y_t\) by iteratively sampling \(y_{t-1}\) until \(t = 1\) according to the following expanded Eq. (\ref{change2}):
\begin{equation}
y_{t-1} = \frac{1}{\sqrt{\alpha_t}} \left( y_t - \frac{1 - \alpha_t}{\sqrt{1 - \bar{\alpha}_t}} \epsilon_{\theta}(y_t, t) \right) + \hat{\beta}_t z,
\label{sampling}
\end{equation}
where \( z \sim \mathcal{N}(0, I) \), and $\hat{\beta}_t$ is usually approximated as $\beta_t$ in practical \cite{DDPM}. Such a sampling process is illustrated in Fig. \ref{sampling_process}.

\subsection{Conditional Diffusion Model}
Similar to the development of GAN, the diffusion model was first proposed with the unconditional generation, and then followed with the conditional generation \cite{text-to-image0}. Unconditional generation is often used to explore the upper limits of model capabilities, while conditional generation is more conducive to applications since it allows for the output to be controlled based on human wishes.

\begin{figure}[t]
\centering
\includegraphics[scale=0.19]{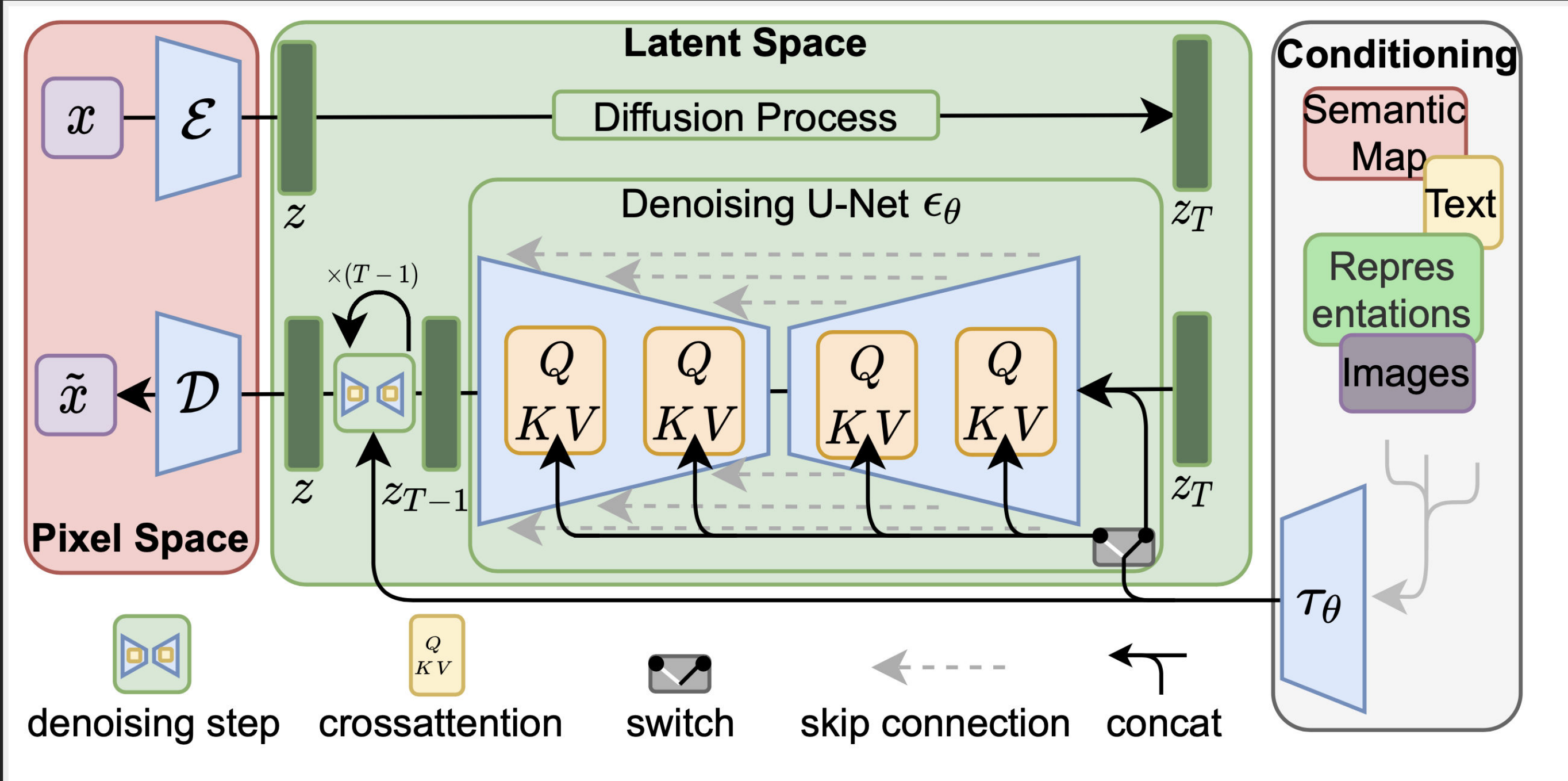}
\caption{\textcolor{black}{The structure of latent diffusion model, where $E$ and $D$ represent the encoder and decoder respectively. The image obtained from \cite{SD}.}}
\label{LDM}
\end{figure}

The first work to introduce the condition in a diffusion model is \cite{CDDMP}, which guides the generation of the diffusion model by adding a classifier to the well-trained diffusion model, so it is also called the \textit{Guided Diffusion Model}. Although this method is less expensive to train, it also increases the inference cost by utilizing classification results to guide the sampling process of the diffusion model. More importantly, it has poor control over details and fails to produce satisfactory required images. As a result, the Google team \cite{CDDPM} decided to adopt a straightforward idea to control the generated results by retraining the DDPM with conditions, and named it as \textit{Classifier-Free Guidance}. 

Formally, given conditional information $c$, the distribution of DDPM that need to be learned is changed as
\begin{gather}
p_{\theta}(x_{t-1} | x_t, c) = \mathcal{N}(x_{t-1}; \mu_{\theta}(x_t, c, t), \tilde{\beta}_t \mathbf{I}), \label{CDDPM0}\\[2mm]
\mu_{\theta}(x_t, c, t)  = \frac{1}{\sqrt{\alpha_t}} \left( x_t - \frac{1 - \alpha_t}{\sqrt{1 - \bar{\alpha}_t}} \epsilon_{\theta}(x_t, c, t) \right).
\label{CDDPM1}
\end{gather}
Correspondingly, the optimization objective (\ref{loss_simp}) and sampling process (\ref{sampling}) are modified into following forms:
\begin{gather}
{{\cal L}}_{con}(\theta ) = \mathbb{E}_{x_0,\epsilon,c,t} \left[ \left\| \epsilon - \epsilon_{\theta} \left( \sqrt{\bar{\alpha}_t}x_0 + \sqrt{1 - \bar{\alpha}_t} \epsilon, c, t \right) \right\|^2 \right], \label{loss_cddpm}\\
y_{t-1} = \frac{1}{\sqrt{\alpha_t}} \left( y_t - \frac{1 - \alpha_t}{\sqrt{1 - \bar{\alpha}_t}} \epsilon_{\theta}(y_t, c, t) \right) + \hat{\beta}_t z.
\label{sampling_cddpm}
\end{gather}

Compared to the \textit{Guided Diffusion Model}, this method is more widely used and is the basis of many attractive models (such as DALL-E2 \cite{AGI4}, Imagen \cite{text-to-image2}, Stable Diffusion \cite{SD} etc.), as well as the theoretical foundation of the conditional diffusion model in RS mentioned below.

\begin{figure*}[t]
\centering
\includegraphics[scale=0.89]{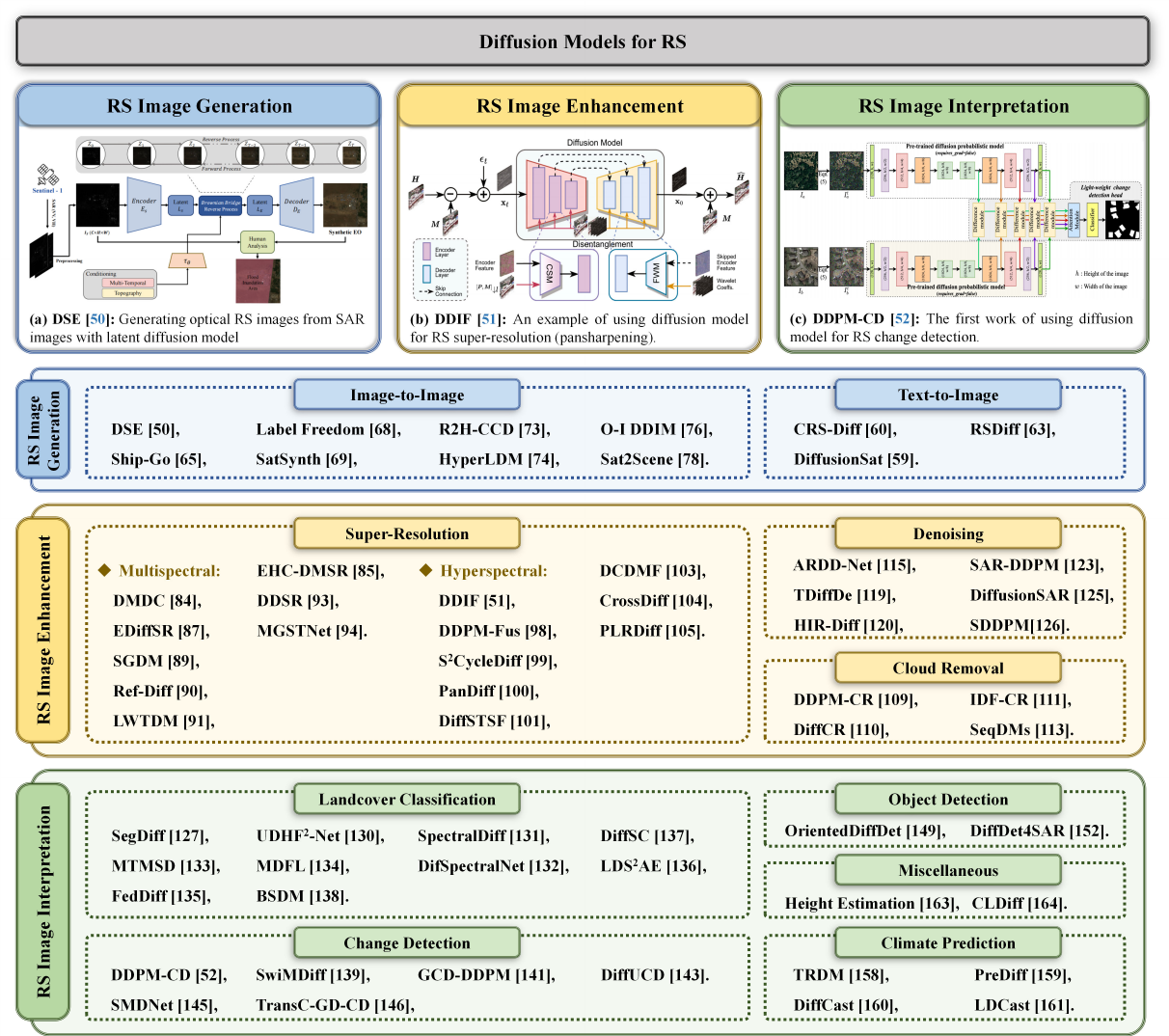}
\caption{\textcolor{black}{The proposed taxonomy of diffusion model applications in RS. The images (a)-(c) are obtained from \cite{52}, \cite{34}, and \cite{37}, respectively.}}
\label{taxonomy}
\end{figure*}

\subsection{\textcolor{black}{Latent Diffusion Model}}
\textcolor{black}{Given the high computational cost of directly operating in the original image pixel space, Rombach \emph{et al.} \cite{SD} proposed to perform the training and inference of diffusion models in a lower dimensional latent space, and named it as the Latent Diffusion Model (LDM).}

\textcolor{black}{As shown in Fig. \ref{LDM}, LDM consists of three parts. Firstly, a pre-trained autoencoder is used to transfer the diffusion model’s working space from the pixel space to the latent space (see the leftmost red box of Fig. \ref{LDM}). Then, the specific diffusion process of LDM is illustrated within the middle green box, where the upper part is the forward diffusion process, that is, adding noise to the feature $z$ to obtain $z_T$. The lower part is the backward diffusion process, where the U-Net with cross-attention supports multi-modal inputs to restore $z_T$ back to $z$. The right part of Fig. \ref{LDM} is a conditional encoder, which encodes various conditions (such as text and images) into a feature vector ${\tau _\theta }$ and feeds it into the U-Net for image generation. The LDM effectively reduces computational cost and speeds up inference without degrading the quality of image synthesis, which is widely used in natural scenes and provides inspiration for the application of diffusion models in RS.}

\section{Applications of Diffusion Models in Remote Sensing}
In this section, we will review and summarize existing related work, all of which involve the use of diffusion models in addressing RS image-related problems. To better organize our review, we categorize these papers according to their applications in RS and provide subdivisions for some common applications, as illustrated in Fig. \ref{taxonomy}. It is important to note that some applications may overlap with each other, but our categorization attempts to align with the core problems addressed by each paper.

\subsection{RS Image Generation}
As one of the most impressive deep generative models, diffusion models are expected to synthesize realistic RS images from existing images or given textual descriptions to support the development of various RS applications, as an alternative to autoregressive models \cite{10287255}. According to the data sources, these image-generation methods can be mainly divided into two categories: text-to-image generation and image-to-image generation.

\subsubsection{Text-to-Image}
Over the past two years, numerous text-to-image diffusion models have come out in computer vision \cite{text-to-image2,SD,text-to-image0}, especially the LDM-based Stable Diffusion (SD) model, which has been widely adopted since its release \cite{SD1,SD2,SD3}. However, its success mainly depends on training with billions of text-image pairs from the internet \cite{SD_dataset}, which makes it difficult to extend to the field of RS since such vast and diverse RS datasets are not readily available, as mentioned in \cite{10287255}. To address this problem, a straightforward idea is to produce trainable RS text-image pairs. Ou \emph{et al.} \cite{7} realized this idea with the help of pre-trained large models. Specifically, they first caption the existing RS images through a vision-language pre-training model to obtain initial textual prompts. Then, refining these prompts with human feedback and GPT-4 to improve semantic accuracy and suitability, successfully enabling the SD model to synthesize the required RS images.
Instead of directly supplementing text prompts, Khanna \emph{et al.} \cite{2} used various numerical information related to satellite images, such as geolocation and sampling time as new prompts of the SD model, which effectively enriches the SD model’s input and enhances its ability to generate high-quality satellite images. Furthermore, Tang \emph{et al.} \cite{70} refined the generation process with SD model by incorporating RS image-related features as control conditions. They treated textual descriptions and numerical information as global control information, and used the depth map, segmentation mask, object boundaries, and other result images obtained through a series of pre-trained networks as local control information. By flexibly selecting control conditions, this approach achieves effective integration of multiple control information, expanding the RS image generation space.

Despite the significant progress made by the improved SD model on optical RS images, researchers recently encountered new challenges when adapting it to other modal RS images \cite{13}. For example, directly using SAR images to fine-tune the SD model will degrade the model's representational ability, resulting in the failure to generate satisfactory SAR images. This is because there are significant differences in capture perspectives and data modalities between SAR and natural images. In view of this, Tian \emph{et al.} \cite{13} proposed to fine-tune the SD model with optical RS images before using SAR images, so as to transition the model from the regular view to the bird's-eye view. Meanwhile, they suggested training the SD model's Low-Rank Adaptation network \cite{Lora}, rather than the whole model, thus ensuring that semantic knowledge learned from natural images can be successfully transferred to the learning process of SAR images.

\begin{figure}[t]
\centering
\includegraphics[scale=0.44]{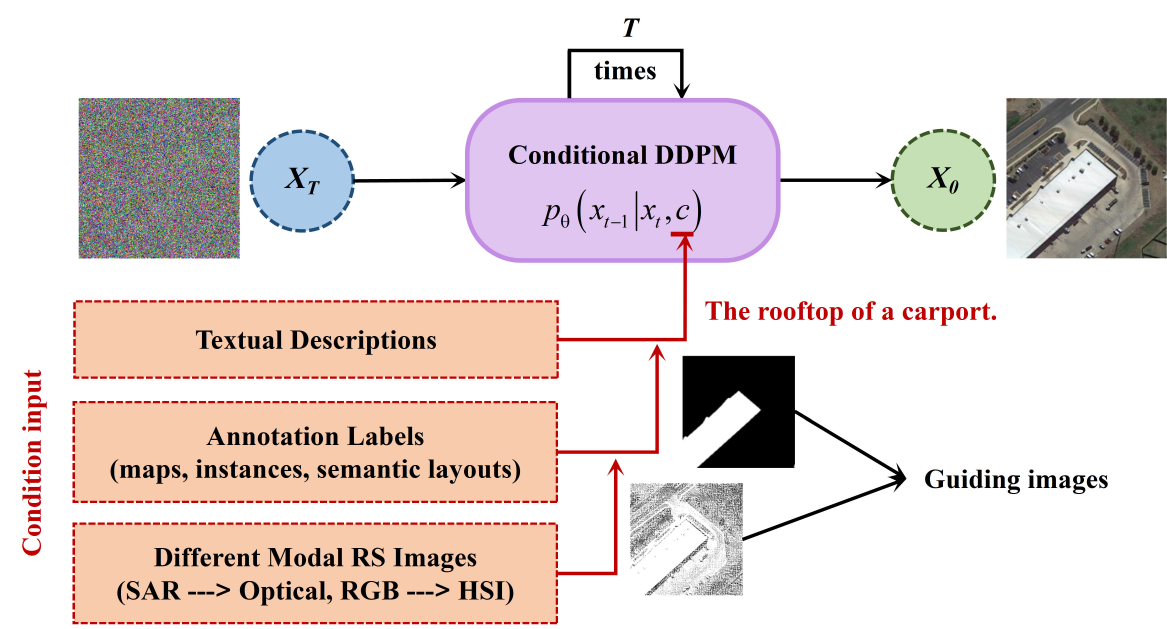}
\caption{Overview of different diffusion model-based methods for RS image generation. Note that also combinations of different condition inputs are possible.}
\label{CDDPM_generation}
\end{figure}

Apart from fine-tuning the SD model, researchers in the RS community have attempted to design a new architecture for RS text-to-image generation with diffusion model \cite{3}. The proposed pipeline consists of two cascaded diffusion models, where the first one is designed to generate low-resolution satellite images from text prompts, and the second one is to increase the resolution of the generated images based on the text descriptions. The benefits of this two-stage generation approach are twofold. On the one hand, separating the generation of low and high-resolution images is advantageous for capturing scene information from global to local perspectives. On the other hand, the low-resolution image generation stage can alleviate the computational burden of generating high-resolution satellite images directly from text descriptions, which is more feasible for practical deployment.

\subsubsection{Image-to-Image}
Compared to the text-to-image generation, image-to-image generation is more popular in the field of RS. In this task, diffusion models are guided by existing images to generate new images. The guiding images can take various forms, such as maps \cite{12}, instances \cite{811}, and semantic layouts \cite{4,9,71,80}. Although these images contain only some specific information, they still present excellent performance in the RS image generation. For example, \cite{12} successfully produced realistic satellite images by training the conditional diffusion model \cite{ControlNet} with maps, even historical maps. \textcolor{black}{\cite{811} used ship instances as inputs for the diffusion model and generated SAR ship images with diverse backgrounds, effectively addressing the issue of insufficient training datasets for SAR ship detection.} \cite{4} not only generated high-quality RS images with the guides of semantic masks, but also addressed the inherent problem of diffusion models requiring long training time for the model convergence. In addition, some researchers are not satisfied with generating high-quality RS images only; they believe that generating image-annotation pairs is more practically valuable. For instance, Zhao \emph{et al.} \cite{71} performed a two-phase fine-tuning on the SD model, achieving the generation from noise to annotations and then from annotations to images. \textcolor{black}{Toker \emph{et al.} \cite{80} extended the standard diffusion model into a joint probability model of images and their corresponding labels, enabling the simultaneous generation of labels and images.}

However, the guiding images used in these methods are essentially annotation labels that require expert knowledge and manual labeling, making them costly to acquire. Given the difficulty in acquiring these images, some works have explored to use multi-modal RS images as the guide images of diffusion models \cite{50,81,5,8}. Different modal RS images have their own strengths and weaknesses. For example, optical RS images are highly visualized and can intuitively reflect surface information, but they are limited by weather conditions and capture time, and are easily obscured by clouds. Conversely, SAR images can be captured in all weather conditions and penetrate clouds and fog, but their imaging process is complex and usually require experts to interpret. In view of these, Bai \emph{et al.} \cite{50} and Shi \emph{et al.} \cite{81} adopted SAR images as the guide images for a diffusion model to generate optical RS images, which gains higher clarity and better structural consistency than those generated by GAN models. Siilarly, hyperspectral images (HSIs) can provide richer spectral-spatial information than multispectral images (e.g., RGB images), even though both belong to optical RS images. However, the acquisition cost of HSIs is much higher than that of multispectral images. Therefore, researchers would like to generate HSIs with the help of easy-obtainable multispectral RS images \cite{5,8}. Unfortunately, using diffusion model to generate HSIs requires matching the input noise dimensionality with the spectral bands of the HSI, resulting in an excessively large noise sampling space that hampers the model's convergence. To address this issue, Zhang \emph{et al.} \cite{5} proposed a spectral folding technology to convert the input HSI into a pseudo-color image before training the diffusion model. \textcolor{black}{Liu \emph{et al.} \cite{8}, inspired by the LDM, used the conditional vector quantized generative adversarial network (VQGAN) \cite{VQGAN} to obtain a latent code space of HSIs, and performed the training and sampling processes of diffusion model within this space.}

Overall, the above image-to-image generation methods are based on the conditional diffusion model \cite{CDDPM}, using the guide image as the condition input to the conditional DDPM, which essentially generates target RS images from noise rather than images directly. To realize the true image-to-image translation, Wang \emph{et al.} \cite{51} employed a straightforward idea: input Inverse Synthetic Aperture Radar (ISAR) images \cite{ISAR1} to the diffusion model during the training phase to force the model to learn the distribution of ISAR images, and then input optical RS images in the testing phase to generate new ISAR images. Seo \emph{et al.} \cite{52} proposed to generate optical RS images from SAR images by sampling noise from the target images rather than using Gaussian noise, efficiently ensuring the consistency of the distribution between the generated image and the target image without using the conditional diffusion model. More recently, Li \emph{et al.} \cite{68} utilized a 3D diffusion model with sparse convolutions to generates texture and colors for the foreground point cloud, comprising buildings and roads, and employed a 2D diffusion model to synthesize the background sky, which enabled the direct generation of 3D scenes solely from satellite imagery, demonstrating a novel application of diffusion models in the RS community. Apart from generating images from one modality to another, researchers have proposed that large satellite images can be generated from patch-level images within the same modality \cite{10}. Specifically, they first adopted self-supervised learning to extract feature embeddings of the input patches, and then used these embeddings as condition input to guide the diffusion model's learning. Notably, they retained the spatial arrangement of each patch in the original image, making it possible to assemble the generated patches into a large and coherent image.

\begin{figure}[t]
\centering
\includegraphics[scale=0.4]{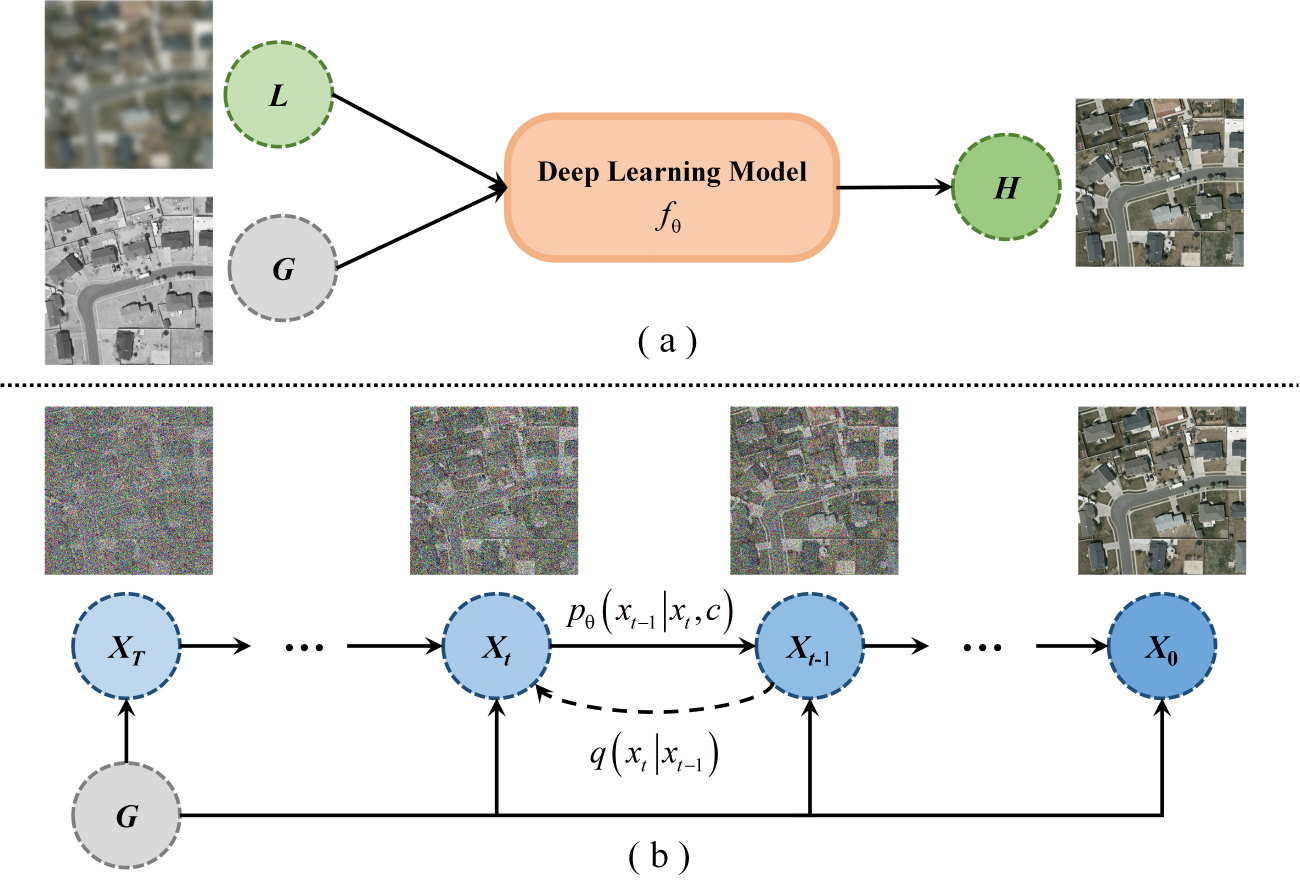}
\caption{Comparison of previous deep learning-based methods and diffusion model-based methods for the RSSR task. (a) The workflow of previous deep learning-based methods, where \textbf{L}, \textbf{G}, \textbf{H} indicate the LRRS image, guided image and HRRS image. (b) The workflow of diffusion model-based methods, where $\textbf{X}_t$ represents the diffused HR image at \textit{t} and \textbf{G} represents the condition input (such as the LR image and its features) for the diffusion model. The figure originally shown in \cite{34}}
\label{HSISR}
\end{figure}

\subsection{RS Image Enhancement}
\subsubsection{\textbf{Super-Resolution}}
Remote sensing super-resolution (RSSR) aims to reconstruct high-resolution (HR) RS images with more details from low-resolution (LR) RS images \cite{RSSR}, which are degraded by imaging equipment, weather condition, or downsampling. Compared to natural images, RS images suffer from much more details loss, making it more challenging to reconstruct the HR images. \textcolor{black}{Before the advent of diffusion models, deep neural network-based RSSR methods, such as Convolutional Neural Networks (CNNs) and Transformers, aimed to find a suitable mapping function to integrate detail information in a single step, as shown in Fig. \ref{HSISR} (a). To achieve this, various techniques were adopted, including self-supervised learning \cite{RSSR1}, adaptive token selection \cite{RSSR2}, and frequency information assistance \cite{RSSR3}. In contrast, diffusion model-based methods fuse the LR image and guided image at each step of the diffusion process (see Fig. \ref{HSISR} (b)), facilitating better integration of different image information and gaining widespread attention in the RS community.}



\paragraph{Multispectral images}
Given that RS images always contain small and dense targets, Liu \emph{et al.} \cite{23} proposed a diffusion model with a detail supplement mechanism for the RSSR task, which requires two-step training to realize. Specifically, the first training aims to improve the model’s capability for reconstructing small objects through randomly masking HR images, and the second training is to complete the super-resolution task by utilizing a conditional diffusion model with LR images as condition input. Although superior performance is achieved, the dual training process in this method is complex and time-consuming. To simplify the training process, Han \emph{et al.} \cite{24} leveraged Transformer \cite{24transformer} and CNN to extract global features and local features from LR images, and used the fused feature images to guide the diffusion model to generate HR images. In this way, the function of two-step training in \cite{23} is successfully realized in one training. Similarly, Xiao \emph{et al.} \cite{22} extracted rich prior knowledge from the original LR images by using stacked residual channel attention blocks \cite{22residual} to guide the optimization of the diffusion model. \textcolor{black}{Apart from obtaining guidance from the LR image, Wang \emph{et al.} \cite{105} and Dong \emph{et al.} \cite{106} found that vector maps and landcover change priors can provide more effective semantic information for diffusion models, helping to ensure the content consistency of the generated HR images.} Furthermore, An \emph{et al.} \cite{21} departed from the commonly used U-Net architecture in diffusion models, implemented an encoder-decoder architecture through parameter-free approaches, and adopted denoising diffusion implicit models (DDIM) \cite{DDIM} to accelerate the sampling process, which significantly improves the efficiency of generating HR images, and is more suitable for diverse RS scenarios.

The above RSSR methods are designed on the assumption that LR images are generated from a fixed degradation model, such as downsampling. However, the blurring in real remote sensing images are complex and varied, which can be modeled as many different degradation models. In view of this, Xu \emph{et al.} \cite{25} proposed to solve this problem with two diffusion models, where the first one is trained as a degradation kernel predictor, so that the predicted degradation kernel and LR image can be used together as conditions in the second diffusion model to generate the HR images. Feng \emph{et al.} \cite{108} achieved the learning of degradation kernel and the reconstruction of HR images within a single diffusion model through the use of the kernel gradient descent module and kernel proximal mapping module.

\paragraph{HSIs}
Although HSIs possess high spectral resolution, their spatial resolution is relatively low \cite{HSIresolution}, which may limit the performance of various applications based on HSIs. To obtain high spatial resolution HSIs, there are two categories methods: pansharpening \cite{PANHSI} and multispectral and hyperspectral image fusion \cite{MultiHSI}. As the name suggests, the former enhances the HSI by injecting the detail information from panchromatic (PAN) images, while the latter leverages multispectral images to help the HSI learn spatial details. For example, Shi \emph{et al.} \cite{32} used the concatenated image of the multispectral and HSI as the condition input for the diffusion model, enabling the model to capture useful information from both image modalities to generate HSIs with high spatial resolution. Qu \emph{et al.} \cite{107} designed a conditional cycle-diffusion framework to allow the spatial SR process and spectral SR process are respectively guided by the multispectral image and the hyperspectral image, thereby generating high-resolution HSIs with complementary spatial and spectral information.

Compared to using multispectral images as the detail guides, researchers are increasingly dedicated to achieve HSI super-resolution with PAN images \cite{30,fusion_new,34,26,31,104,29}. Instead of crudely concatenating the PAN and HSI images directly, Meng \emph{et al.} \cite{30} proposed a Modal Intercalibration Module to enhance and extract features from both images, where the enhanced features is used as condition input to the diffusion model. Cao \emph{et al.} \cite{34} believed that the unique information in different image modalities should not be blended for processing. Thus, they designed two conditional modulation modules to extract coarse-grained style information and fine-grained frequency information respectively as the condition inputs. Still for fully utilizing the unique information of different modalities, \textcolor{black}{Li \emph{et al.} \cite{31} proposed a dual conditional diffusion models-based pansharpening network, which takes HSI and PAN as independent condition inputs to learn the spectral features and spatial texture respectively. Notably, this network structure is designed based on the LDM, which not only achieves excellent SR performance, but also greatly reduces the computational cost.} \cite{29} also performs the sampling process in a low-dimensional space. Based on the assumption that the HRHSI can be decomposed into the product of two low-rank tensors, this method first computes one of the low-rank tensors with the LRHSI. Then, this tensor is taken as the condition along with the LRHSI and PAN, input into a pre-trained RS diffusion model \cite{37} for another low-rank tensor. This method, unlike the above-mentioned methods, is a completely unsupervised deep learning method that does not require the involvement of HRHSI in any process, providing feasibility for its application in practice.

\subsubsection{\textbf{Cloud Removal}}
In many cases, optical RS images will be partially obscured by clouds since the view of imaging equipment is limited. Cloud removal, in essence, is to reconstruct the areas corrupted by clouds, which means to generate content that is consistent with the surrounding environment to fill the missing areas of the image \cite{44}. Therefore, it is suitable to use diffusion model for this task, since it can better control the generated content.

The control condition for cloudy image reconstruction comes in various forms. For example, Czerkawski \emph{et al.} \cite{54} adopted text prompts and edge information \cite{edge} as the guiding conditions, together with the input cloudy image, cloud mask, and diffused cloud-free image to control the generation process of SD \cite{SD}. Jing \emph{et al.} \cite{42} input both SAR image and cloudy optical RS image into the diffusion model for feature extraction, effectively enhancing the cloud removal results with the help of cloud-unaffected SAR image. Zou \emph{et al.} \cite{43} first extracted features from the cloudy image and noise level, and then input the extracted spatial and temporal features into the diffusion model as control conditions. Instead of using more controlled conditions, they trained the model in a supervised manner with cloud-free images to further improve the quality of the reconstructed images. \textcolor{black}{Wang \emph{et al.} \cite{94} and Sui \emph{et al.} \cite{93} also used cloud-free images, but in different ways. Wang \emph{et al.}\cite{94} first generated the cloud-free image through a pre-trained diffusion model, and then incorporated it into the cloud removal diffusion process as the control condition. In contrast, Sui \emph{et al.} \cite{93} directly used the cloud-free image as input of the diffusion model.}
Moreover, Zhao \emph{et al.} \cite{41} integrated different images from multi-modalities and multi-time into a sequence input, and utilized a two-branch diffusion model to extract scene content from the optical RS and SAR images, respectively. Unlike the previous works, this method does not require the participation of cloud-free images and can handle image sequences of any length, offering greater flexibility and practical value.

\subsubsection{\textbf{Denoising}}
Due to the inherent constraints of imaging technology and environmental conditions, RS images are always accompanied by various noise \cite{RSdenoise}. In essence, the learning process of the diffusion model is equal to the denoising process \cite{DDPM}, which makes it possible to achieve superior denoising performance in the context of RS. Different modalities of RS images confront different noise challenges. For example, optical RS images are more susceptible to blurring caused by atmospheric scattering and absorption \cite{45,denoise242}, as well as the noise from changing lighting conditions \cite{62}. HSIs need to consider the uneven distribution of noise over the spectral dimensions \cite{16} and the correlation of noise between different bands \cite{15,denoiseHSI24}. Fortunately, these noises can be effectively removed through diffusion models. Huang \emph{et al.} \cite{45} proposed to crop the noisy RS image into small regions and rearrange them in a cyclic shift manner before fed into the diffusion model, so as to achieve finer local denoising as well as artifact elimination. He \emph{et al.} \cite{15} proposed a truncated diffusion model that starts denoising from the intermediate step of the diffusion process, instead of a pure noise, to avoid the destruction of the inherent effective information in the HSI. Moreover, Yu \emph{et al.} \cite{20} simulated the harsh imaging conditions of RS satellites by adding various attack disturbances to input images, enhancing the diffusion model's ability to counteract the system noise.

\begin{figure}[t]
\centering
\includegraphics[scale=0.42]{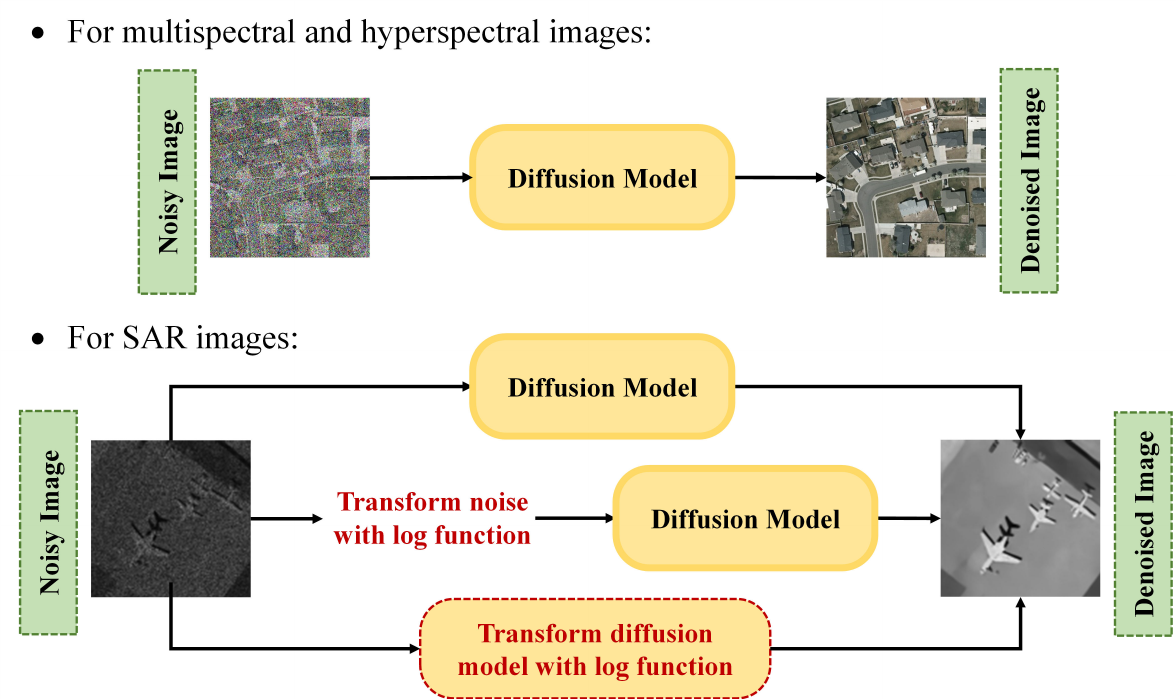}
\caption{Overview of diffusion model-based methods for RS image denoising. Since multispectral and hyperspectral images typically exhibit additive noise, diffusion models can be directly employed for denoising. In contrast, SAR images are corrupted by the multiplicative noise. Consequently, there are three approaches to address this problem: directly utilizing the diffusion model, transforming the multiplicative noise into additive noise, as well as transforming the diffusion model to be fit for removing multiplicative noise.}
\label{Denoise}
\end{figure}

SAR images, as another modality of RS images, are usually contaminated by a multiplicative noise, speckle \cite{SARnoise}. Unlike additive noise, the degradation caused by speckle varies across different areas in one image. Therefore, speckle noise significantly affects the disparity and interdependence between pixels, causing severe damage to the image. To eliminate this particular noise, Perera \emph{et al.} \cite{18} proposed to use the speckled SAR image, along with the Gaussian noised SAR image that conforms to the standard diffusion model for denoising training. However, this method of introducing the synthetic noise images may not be able to accurately and reasonably simulate the actual SAR images, causing suboptimal denoising performance. \textcolor{black}{Thus, Xiao \emph{et al.} \cite{17} and Ma \emph{et al.} \cite{78denoise_new0} respectively adopted the log function and Yeo-Johnson transformation to convert the multiplicative noise in SAR into additive noise, enabling the transformed SAR images to match a standard diffusion model for independent training.} Further advancing this methodology, Guha \emph{et al.} \cite{19} integrated the log operation into the derivation process of the diffusion model and obtained a one-step denoising network that can directly address the multiplicative noise.
\textcolor{black}{The overall procedure of denoising different modalities RS images with diffusion models is illustrated in Fig. \ref{Denoise}}

\begin{figure}[t]
\centering
\includegraphics[scale=0.42]{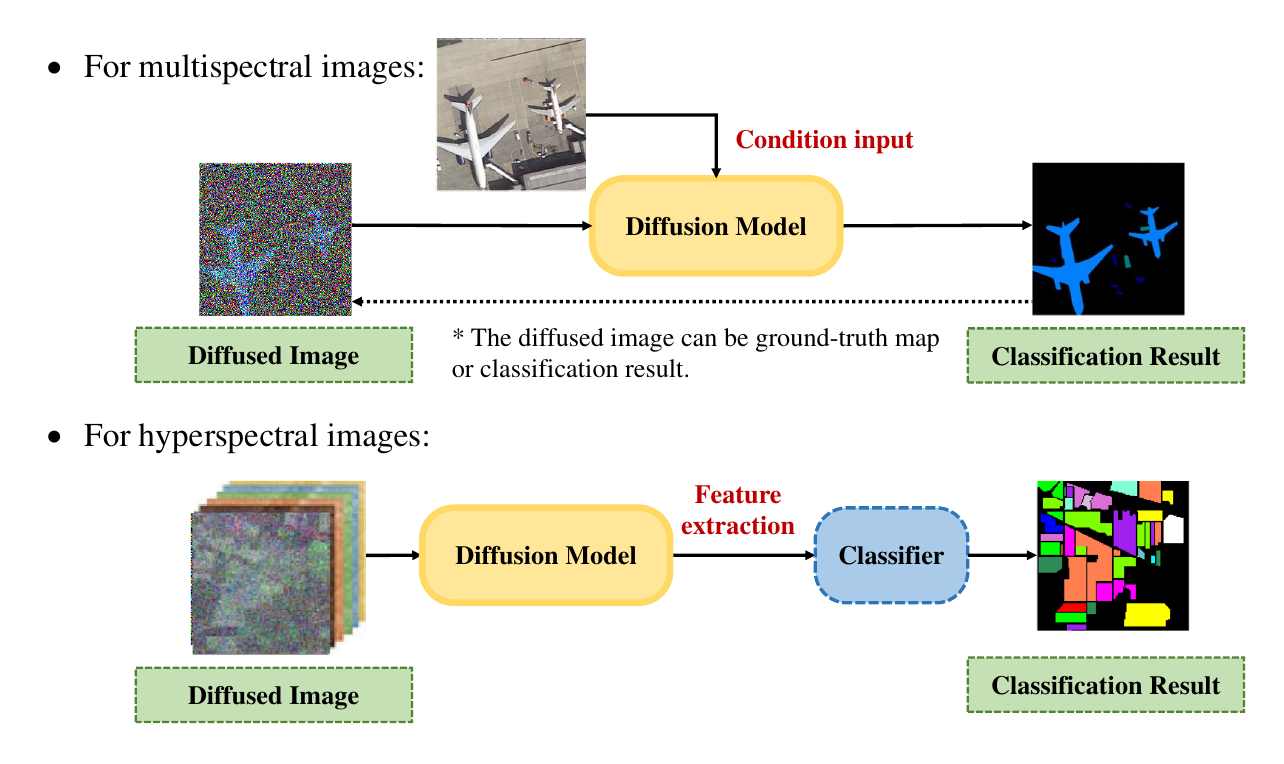}
\caption{\textcolor{black}{Overview of diffusion model-based methods for landcover classification. For multispectral images, the ground-truth map is typically used as the diffused image, with the original RS image serving as the condition. The dashed line indicates that the classification results predicted by the network can also be used as the diffused image for the next input. As for HSIs, the prevailing approaches involve using the diffusion model as a feature extractor, after which the extracted features are fed into a classifier for classification.}}
\label{classfier}
\end{figure}

\subsection{RS Image Interpretation}
\subsubsection{\textbf{Landcover Classification}}
As one of the most common applications in the RS community, landcover classification aims to assign each pixel to a specific class, such as buildings or grass, to obtain useful landcover information. However, the RS images always contain diverse and complex scenes, increasing the difficulty of accurate classification. Given that the diffusion model can learn and simulate complex data distributions better than other deep learning models, researchers are trying to use it for the RS landcover classification. 

The first application of the diffusion model to this task is presented in \cite{49}, named SegDiff. Based on the conditional diffusion model \cite{CDDMP}, SegDiff takes the manually annotated ground-truth map and the original RS image as the diffused image and condition respectively, and decouples the commonly used U-Net architecture by adding two separate encoders for extracting features from both diffused and guiding images. Besides, it averages the results of multiple sampling for the final classification result to improve the stability and overall accuracy. This method was later tested by Ayala \emph{et al.} \cite{48} on wider RS datasets, validating the effectiveness and development potential of the diffusion model in the RS landcover classification. Instead of using the ground-truth map as the diffused image, Kolbeinsson \emph{et al.} \cite{60} diffused the classification prediction result of the network from the previous step and input it along with the conditioning RS image into the diffusion model for next classification prediction. Notably, the parameters of their diffusion model are optimized not only through the MSE of predicted noise but also by minimizing the difference between the prediction result and ground-truth map at each step. \textcolor{black}{Similarly, Zhang \emph{et al.} \cite{83} further subdivided the initial classification results into the certain and uncertain regions, and performed diffusion process on the uncertain regions to improve the robustness of edge extraction.}

Compared to multispectral images, HSIs exhibit a more complex data distribution, posing a greater challenge for the application of conventional deep learning models in landcover classification. Fortunately, the diffusion model can better capture the spectral-spatial joint features of HSIs, facilitating the improvement of classification accuracy \cite{38,85}. Based on this fact, Zhou \emph{et al.} \cite{40} constructed a timestep-wise feature bank by utilizing the temporal information of the diffusion model, and proposed a dynamic fusion module to integrate spectral-spatial features with temporal features, making it possible to obtain sufficient image information before the classification. Li \emph{et al.} \cite{39} designed a dual-branch diffusion model for feature extraction from HSI and LiDAR images separately, achieving information complementarity between different modal RS images and enhancing the distinguishability among pixels, which also demonstrated superior performance in a multi-client RS task \cite{64}. Qu \emph{et al.} \cite{classHSI24} further set the encoder of the dual-branch diffusion model to operate in parameter-sharing mode to ensure the extraction of shared features from multimodal RS images.
In addition, Chen \emph{et al.} \cite{57} utilized the diffusion model to assist deep subspace construction, achieving excellent HSI classification performance in an unsupervised manner. Different from the aforementioned methods, Ma \emph{et al.} \cite{55} used the diffusion model to directly classify the HSI pixels into background and anomaly targets, rather than as a feature extractor. Specifically, they adopted a diffusion model to learn the background distribution based on the fact that the distribution of the background obeys a mixed Gaussian. Thus, the background is removed as noise during the inference process, effectively retaining the anomalous pixels of interest. \textcolor{black}{Fig. \ref{classfier} summarizes the differences in using diffusion models for landcover classification in multispectral and hyperspectral images.}

\begin{figure}[t]
\centering
\includegraphics[scale=0.44]{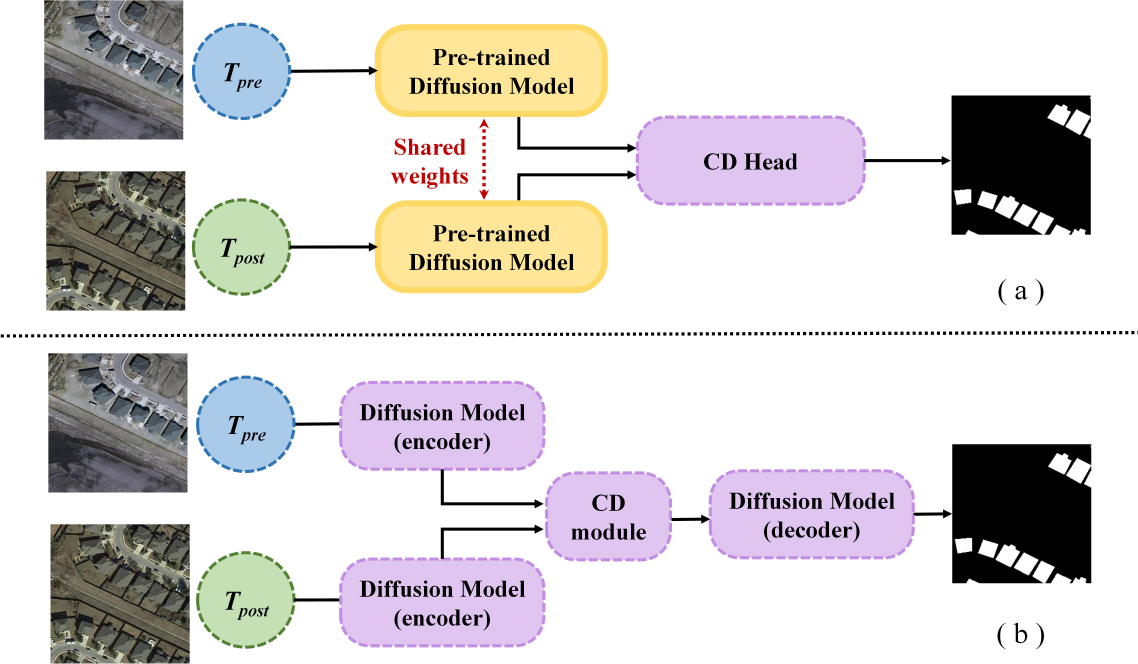}
\caption{Comparison of (a) two-step and (b) end-to-end diffusion model-based CD methods. The two-step methods separate the training process of diffusion model and CD head, where the diffusion models (in yellow) are pre-trained by millions of RS images and only the CD head is optimized by the CD dataset. In contrast, the end-to-end method enables the joint optimization of diffusion models and CD module. The figure originally shown in \cite{35}.}
\label{CD_frame}
\end{figure}

\subsubsection{\textbf{Change Detection}}
RS change detection (CD) aims to identify the differences between two images taken at different times of the same area \cite{CD}, thereby providing support for environmental monitoring, and natural disaster assessment. Considering the exceptional performance of diffusion models in handling image details, such as textures and edges, which are crucial for distinguishing changes, some researchers have explored diffusion model-based CD methods. Bandara \emph{et al.} \cite{37} employed a pre-trained diffusion model to extract multi-scale features of different temporal images for the CD module training, where the pre-training was accomplished by millions of free and unlabeled RS images to capture the key semantic. Tian \emph{et al.} \cite{65} integrated a diffusion model into the contrastive learning framework \cite{moco} to capture fine-grained information in the RS images, successfully extracting features with clearer boundaries and richer texture details for the CD task. Additionally, Zhang \emph{et al.} \cite{36} adopted the Transformer \cite{U-ViT} as the backbone for the diffusion model to extract spectral-spatial features from HSIs captured at different times.

In essence, all of the above methods use the diffusion model as a feature extractor trained separately from the CD task, which results in the generated features not being fully suitable for CD and overlooks the potential benefits of gradual learning and controllability provided by the diffusion model. To address these issues, Wen \emph{et al.} \cite{35} and Jia \emph{et al.} \cite{88} proposed an end-to-end diffusion model for CD, where the diffused ground-truth map is input into the network along with the pre- and post-change images, and the difference between two temporal images is used as the condition to guide the direction of detection. \textcolor{black}{The comparison between separated and end-to-end diffusion model-based CD methods is displayed in Fig. \ref{CD_frame}.
Recently, Wen \emph{et al.} \cite{89} further improved this framework. Instead of direct subtraction and addition, they used neural network modules to obtain the guidance conditions and incorporate them into the diffusion model. This approach effectively bridged the gap between change features and diffusion noise, improving the accuracy of CD results.}

\subsubsection{\textbf{Object Detection}}
Different from the pixel-wise landcover classification, RS object detection uses bounding boxes to locate the instances of a certain class (such as plane, vehicle, or ship) in the images. One of the most challenging issues in RS object detection is the lack of sufficient training data \cite{GAN2}. This scarcity is due to the long-distance photography of RS images, which often results in the object of interest being small and sparsely distributed across different regions in the images. Thus, augmenting the object of interest with the diffusion model has become an effective solution \cite{63}. Specifically, it works by training the SD model with object patches cropped from the available object detection training set, and then pasting the generated object patches back onto the real backgrounds. \textcolor{black}{Recently, Wang \emph{et al.} \cite{91} were inspired by \cite{diffusiondet} and introduced a novel RS object detection diffusion model. This method ingeniously integrates object detection with diffusion model, converting the task of pinpointing an object’s exact position into a denoising process. Specifically, it gradually refined random noise boxes into the accurate target bounding box and showed favorable performance on the DOTA dataset \cite{dota}. Later, Zhou \emph{et al.} \cite{92} applied it to SAR images and introduced a scattering feature enhancement module to further improve the separability of background and SAR object.}

\begin{figure*}[t]
\centering
\includegraphics[scale=0.46]{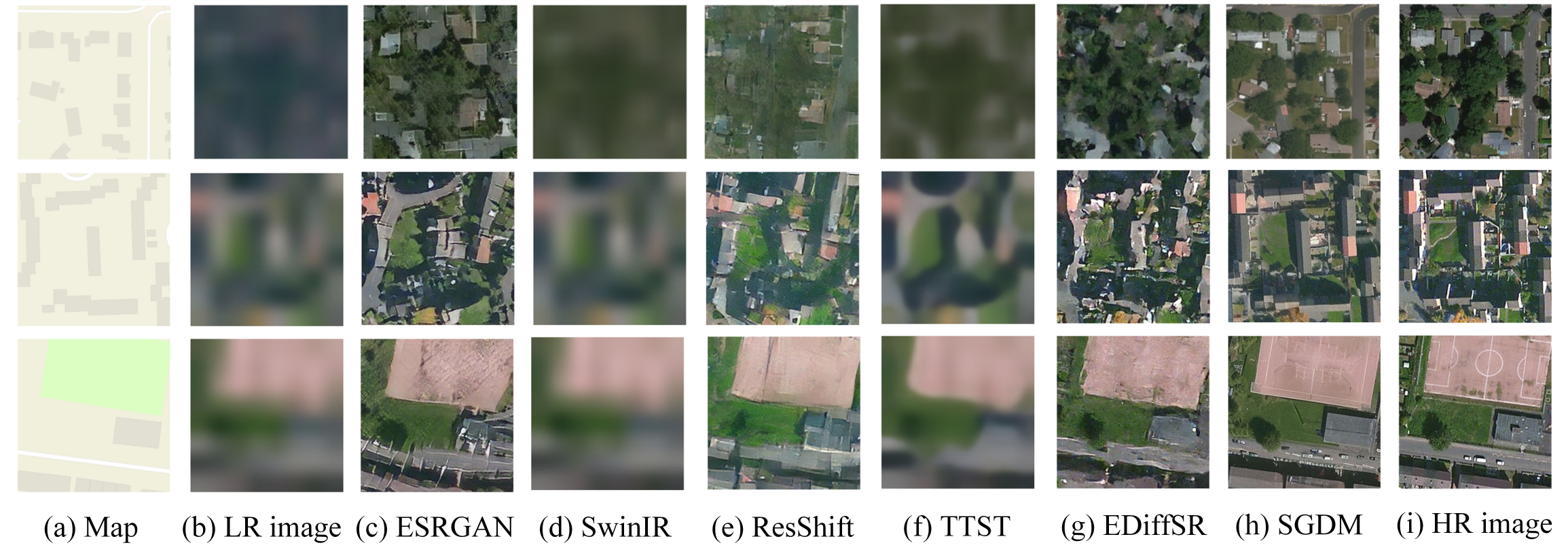}
\caption{\textcolor{black}{Comparison of different super-resolution methods on RS images: (a) Map (i.e., Guided image); (b) LR image; (c) ESRGAN \cite{esrgan}; (d) SwinIR \cite{swinir}; (e) ResShift \cite{resshift}; (f) TTST \cite{RSSR2}; (g) EDiffSR \cite{22}; (h) SGDM \cite{105}; (i) HR image. The visual results are retrieved from \cite{105}.}}
\label{SR_result}
\end{figure*}

\subsubsection{\textbf{Climate Prediction}}
Climate prediction is another application for the diffusion model in the field of RS, which is a complex systematic project that requires the integration of multiple variables, including cloud amount, cyclone distribution, and water vapor. Therefore, its solution often consists of multiple diffusion models, enabling fine-grained, stepwise processing of these high-dimensional data. For example, Hatanaka \emph{et al.} \cite{66} utilized two cascaded score-based diffusion models to generate high-resolution cloud cover images from coarse-resolution atmospheric variables. Nath \emph{et al.} \cite{61} cascaded three independently trained diffusion models to generate future satellite imagery, increase the resolution, and predict precipitation. \textcolor{black}{Ling \emph{et al.} \cite{99} used 3D and 2D diffusion models to learn the spatiotemporal features, improving the accuracy of rainfall prediction. More recently, Gao \emph{et al.} \cite{100} and Yu \emph{et al.} \cite{101} proposed PreDiff and DiffCast to achieve precipitation prediction with only one diffusion model.
In addition, Leinonen \emph{et al.} \cite{67} addressed the issue of high computational costs by adopting the LDM framework, running the diffusion process within a latent variable space mapped by a 3D VAE network.
Zhao \emph{et al.} \cite{102} reduced the sampling time of the diffusion model by implementing large-step adversarial mapping in the backward diffusion process, achieving real-time forecasting.}
These methods all demonstrate that diffusion models can better capture the complex spatio-temporal relationships and become a powerful tool in climate prediction.




\subsubsection{\textbf{Miscellaneous Tasks}}
Apart from the classic RS interpretation tasks reviewed above, there are some other tasks that may not fall into the above categories \cite{58,90,69}.

One such task is height estimation, which aims at providing pixel-wise height information of surface features (e.g., buildings, trees, terrain, etc.) to generate 3D models of surface scenes. Diffusion model has been reported to be a promising solution for surface estimation \cite{58}. Unlike traditional methods that require multi-view geospatial imagery or LiDAR point clouds, it produces accurate height estimates only with single-view optical RS images. 

\textcolor{black}{The other task is cloud detection, which aims to separate cloud and clear pixels in RS images. The outcome of this detection enables more efficient processing for the RS images by disregarding the cloud-covered areas. However, detecting thin clouds is particularly challenging since they often appear gray and semitransparent, blending with the background. Fortunately, recent work \cite{90} has demonstrated that diffusion models can effectively address this problem by capturing intraclass variations of clouds, generating a more comprehensive cloud mask.}

Additionally, diffusion model also shows superior performance over CNNs in the task of anomaly detection in satellite videos (such as wildfire detection) \cite{69}. The past frames serve as the condition input for the diffusion model, enabling it to learn the data distribution of normal frames and generate high-quality data that closely resemble real images. Consequently, when an anomalous frame (such as one with wildfire appearance) is input, the model outputs a significantly higher anomaly score. This means that the diffusion model can detect small wildfires promptly, preventing widespread fire outbreaks, which is as opposed to CNN-based methods that usually require the fire to reach a certain visual extent to be effective.

\begin{table}[t]
\centering
\caption{\textcolor{black}{Quantitative Comparison of Different Remote Sensing Super-Resolution Methods}}
\label{SR_table} 
\renewcommand{\arraystretch}{1.3}
\setlength\tabcolsep{2.4mm}{
\begin{tabular}{ccccc}
\hline
\multirow{2}{*}{\textcolor{black}{Methods}}                      & \multicolumn{4}{c}{\textcolor{black}{Metrics}}      \\ \cline{2-5} 
            & \textcolor{black}{LPIPS $\downarrow$}   & \textcolor{black}{FID $\downarrow$}  & \textcolor{black}{MUSIQ $\uparrow$}     & \textcolor{black}{CLIPIQA $\uparrow$}  \\ \hline
\textcolor{black}{ESRGAN \cite{esrgan}}    & \textcolor{black}{0.507} & \textcolor{black}{84.4} & \textcolor{black}{41.18}& \textcolor{black}{0.534}\\ 
\textcolor{black}{SwinIR \cite{swinir}}  & \textcolor{black}{0.903} & \textcolor{black}{390.5}& \textcolor{black}{20.25}& \textcolor{black}{0.263}\\ 
\textcolor{black}{ResShift \cite{resshift}}   & \textcolor{black}{0.540} & \textcolor{black}{126.3}& \textcolor{black}{40.86}& \textcolor{black}{0.427}\\ 
\textcolor{black}{TTST \cite{RSSR2}} & \textcolor{black}{0.875} & \textcolor{black}{327.5}& \textcolor{black}{18.44}& \textcolor{black}{0.251}\\
\textcolor{black}{EDiffSR \cite{22}}            & \textcolor{black}{0.529}& \textcolor{black}{132.2}& \textcolor{black}{40.38}& \textcolor{black}{0.449}\\ \hline 
\textcolor{black}{SGDM \cite{105}}            & \textcolor{black}{0.516} & \textcolor{black}{40.7}& \textcolor{black}{44.61}& \textcolor{black}{0.615}\\ \hline
\multicolumn{5}{l}{\textcolor{black}{*The results are the average of \cite{105}.}}
\end{tabular}
}
\end{table}

\begin{figure*}[t]
\centering
\includegraphics[scale=0.4]{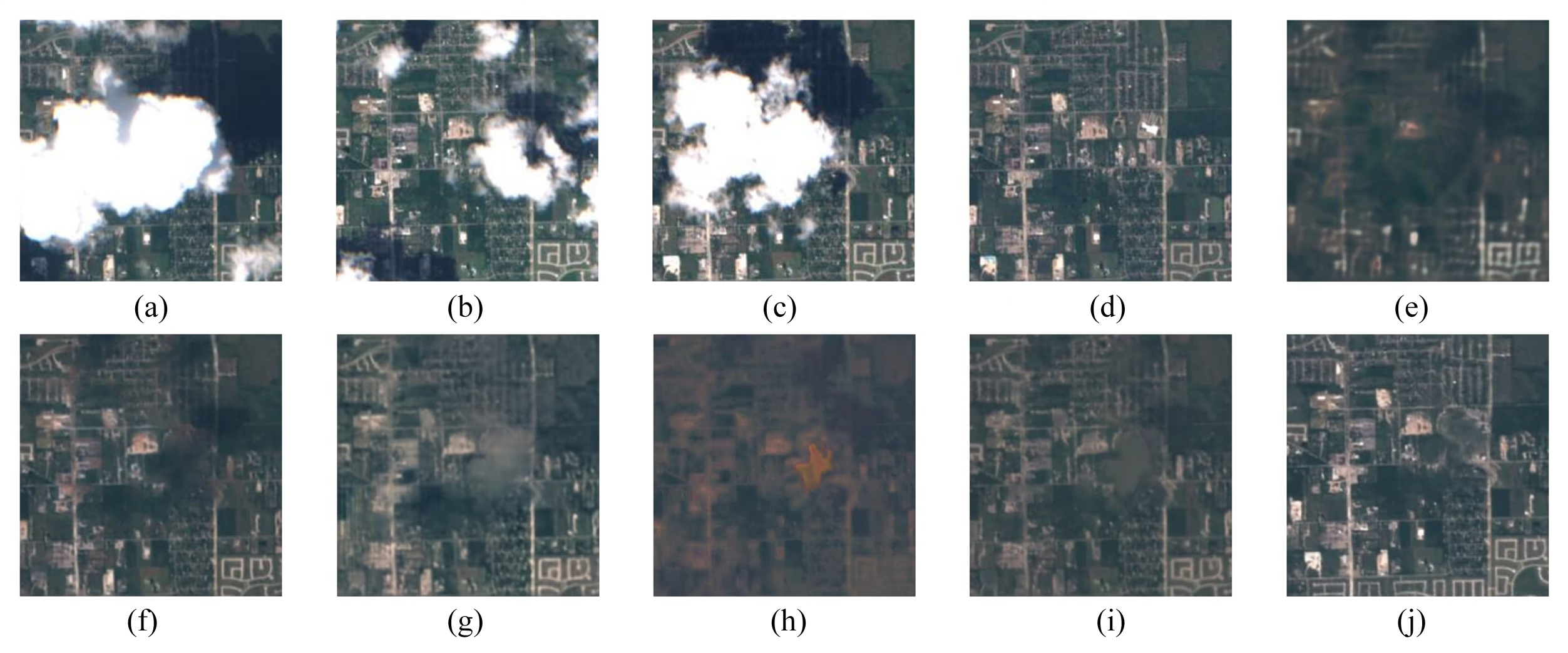}
\caption{Comparison of different cloud removal methods on the Sen2-MTC-Old dataset: (a) Cloudy Image T1. (b) Cloudy Image T2. (c) Cloudy Image T3. (d) Ground-Truth. (e) STNet \cite{stnet}. (f) DSen2-CR \cite{DSen2}. (g) PMAA \cite{PMAA}. (h) UnCRtainTS \cite{UnCRtainTS}. (i) DDPM-CR \cite{42}. (j) DiffCR \cite{43}. The visual results are retrieved from \cite{43}.}
\label{cloud}
\end{figure*}

\begin{figure*}[t]
\centering
\includegraphics[scale=0.7]{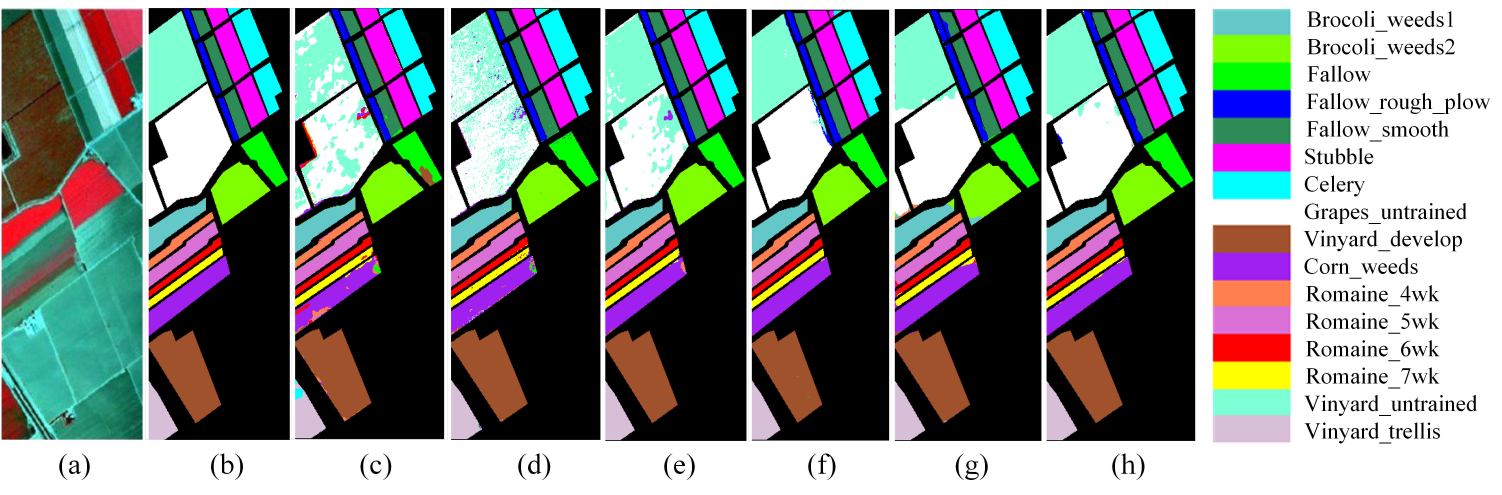}
\caption{Comparison of different HSI classification methods on the Salinas dataset: (a) Pseudo-Color Image. (b) Ground-Truth. (c) SF \cite{SF}. (d) miniGCN \cite{GCN}. (e) SSFTT \cite{SSFTT}. (f) DMVL \cite{DMVL}. (g) SSGRN \cite{SSGRN}. (h) SpectralDiff \cite{38}. The visual results are retrieved from \cite{38}.}
\label{HSI_class}
\end{figure*}

\subsection{Experimental Evaluation}
To effectively illustrate the superiority of diffusion models in processing RS images, we take the experimental results of super-resolution, cloud removal, landcover classification, and change detection as examples in this section to evaluate the performance of the diffusion model and other existing techniques through visual results and quantitative indicators.

\begin{table}[t]
\centering
\caption{Quantitative Comparison of Different Cloud Removal Methods on The Sen2-MTC-Old Dataset}
\label{cloud_table} 
\renewcommand{\arraystretch}{1.3}
\setlength\tabcolsep{2.4mm}{
\begin{tabular}{ccccc}
\hline
\multirow{2}{*}{Methods}                      & \multicolumn{4}{c}{Metrics}      \\ \cline{2-5} 
                                              & PSNR $\uparrow$   & SSIM $\uparrow$  & FID $\downarrow$     & LPIPS $\downarrow$ \\ \hline
STNet \cite{stnet}           & 26.321 & 0.834 & 146.057 & 0.438 \\ 
DSen2-CR \cite{DSen2}        & 26.967 & 0.855 & 123.382 & 0.330 \\ 
PMAA \cite{PMAA}             & 27.377 & 0.861 & 120.393 & 0.367 \\ 
UnCRtainTS \cite{UnCRtainTS} & 26.417 & 0.837 & 130.875 & 0.400 \\ \hline
DDPM-CR \cite{42}            & 27.060 & 0.854 & 110.919 & 0.320 \\ 
DiffCR \cite{43}             & 29.112 & 0.886 & 89.845  & 0.258 \\ \hline
\multicolumn{5}{l}{*The results are retrieved from \cite{43}.}
\end{tabular}
}
\end{table}

\begin{table}[t]
\centering
\caption{Quantitative Comparison of Different HSI Classification Methods on The Salinas Dataset}
\label{classification_table} 
\renewcommand{\arraystretch}{1.3}
\setlength\tabcolsep{4.2mm}{
\begin{tabular}{cccc}
\hline
\multirow{2}{*}{Methods}                 & \multicolumn{3}{c}{Metrics}                 \\ \cline{2-4} 
                                         & OA (\%) & AA (\%) & $\kappa$ (\%) \\ \hline
SF \cite{SF}            & 88.248 & 93.262 & 86.973                    \\
miniGCN \cite{GCN}      & 88.181 & 94.297 & 86.823                    \\
SSFTT \cite{SSFTT}      & 95.789 & 98.272 & 95.322                    \\
DMVL \cite{DMVL}        & 97.005 & 95.853 & 96.668                    \\
SSGRN \cite{SSGRN}      & 96.539 & 96.354 & 96.144                    \\ \hline
SpectralDiff \cite{38}  & 98.971 & 99.465 & 98.854                    \\ \hline
\multicolumn{4}{l}{*The results are retrieved from \cite{38}.}
\end{tabular}
}
\end{table}

\textcolor{black}{As shown in Fig. \ref{SR_result}, the visual results of the transformer-based methods (SwinIR \cite{swinir} and TTST \cite{RSSR2}) are significantly inferior to those based on the diffusion models (ResShift \cite{resshift}, EDiffSR \cite{22} and SGDM \cite{105}). Meanwhile, Table \ref{SR_table} provides the quantitative results of these six comparison methods by using Learned Perceptual Image Patch Similarity (LPIPS) \cite{LPIPS}, Frechet Inception Distance (FID) \cite{FID}, multiscale image quality transformer (MUSIQ) \cite{musiq}, and CLIP Image Quality Assessment (CLIPIQA) \cite{clipiqa}. Although diffusion model-based SR methods lag slightly behind ESRGAN on the LPIPS indicator, they perform well on the other three indicators. Notably, we use two types of RS images, synthetic and real-world data, for our experiments. Thus, the results presented in Table \ref{SR_table} are the average of the outcomes from these two types of images.}

\begin{figure*}[t]
\centering
\includegraphics[scale=0.39]{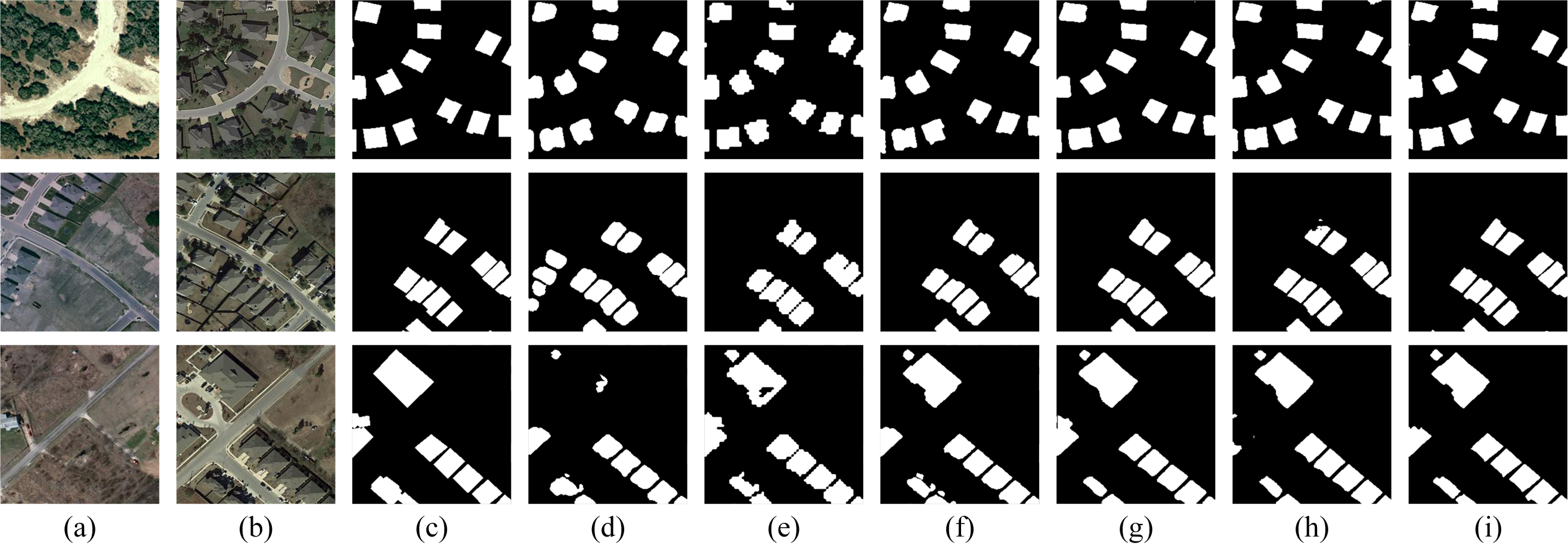}
\caption{Comparison of different change detection methods on the LEVIR dataset: (a) Pre-change Image. (b) Post-change Image. (c) Ground-Truth. (d) FC-SD \cite{fcsd}. (e) STANet \cite{stanet}. (f) SNUNet \cite{snunet}. (g) BIT \cite{BIT}. (h) ChangeFormer \cite{changeformer}. (i) DDPM-CD \cite{37}.}
\label{CD_comparsion}
\end{figure*}

Fig. \ref{cloud} displays the visual results of six different cloud removal methods on multi-temporal optical satellite (Sentinel-2) images \cite{cloud_dataset}, where the first three cloudy images are taken at different times from the same location. Diffusion model-based methods DDPM-CR \cite{42} and DiffCR \cite{43} have successfully removed clouds without leaving excessive artifacts, restoring the RS image with detailed information. This observation is also confirmed by the quantitative indicators. As shown in Table \ref{cloud_table}, Peak Signal-to-Noise Ratio (PSNR), Structural Similarity Index Measure (SSIM) \cite{SSIM}, LPIPS and FID are used in to evaluate the quality of cloud-free images generated by the comparison methods. It is evident that DiffCR achieves the best performance across all four indicators, with DDPM-CR, another diffusion model-based method, also ranking second on the FID and LPIPS indicators. This demonstrates that using diffusion models for cloud removal is highly competitive.

As illustrated in Fig. \ref{HSI_class}, the diffusion model-based method SpectralDiff \cite{38} has significantly better classification results than SF \cite{SF}, miniGCN \cite{GCN}, and SSFTT \cite{SSFTT} on the hyperspectral dataset Salinas. Although DMVL \cite{DMVL} and SSGRN \cite{SSGRN} show comparable performance to SpectralDiff across most classes, they are not as accurate as SpectralDiff in assigning pixels at the boundaries of different classes. In addition to the visualized results, Table \ref{classification_table} presents the quantitative results of these methods on overall accuracy (OA), average accuracy (AA), and Kappa coefficient, where SpectralDiff ranks first.

As for the CD task, we select five comparised methods based on different deep learning models along with a diffusion model-based method, DDPM-CD \cite{37}, for experimentation on the LEVIR dataset \cite{stanet}. 
\textcolor{black}{In this task, we use F1 score (F1), overall accuracy (OA), and mean Intersection over Union (mIoU) including both change and unchanged regions as evaluation indicators. The values of these metrics are listed in Table \ref{CD_table}, and the visual results are displayed in Fig. \ref{CD_comparsion}.} Both qualitative analysis and quantitative comparisons clearly demonstrate that the diffusion model-based CD method is significantly superior to other methods.

\begin{table}[t]
\centering
\caption{Quantitative Comparison of Different Change Detection Methods on The LEVIR Dataset}
\label{CD_table} 
\renewcommand{\arraystretch}{1.3}
\setlength\tabcolsep{4.2mm}{
\begin{tabular}{cccc}
\hline
\multirow{2}{*}{Methods}                            & \multicolumn{3}{c}{Metrics}      \\ \cline{2-4} 
                                                    & F1(\%)    & \textcolor{black}{mIoU(\%)}    & OA(\%)  \\ \hline
FC-SD \cite{fcsd}                  & 86.31     & 75.92      & 98.67   \\
STANet \cite{stanet}               & 87.26     & 77.40      & 98.66   \\
SNUNet \cite{snunet}               & 88.16     & 78.83      & 98.82   \\
BIT \cite{BIT}                     & 89.31     & 80.68      & 98.92   \\
ChangeFormer \cite{changeformer}   & 90.40     & 82.48      & 99.04   \\ \hline
DDPM-CD \cite{37}                  & 90.91     & 83.35      & 99.09   \\ \hline
\multicolumn{4}{l}{*The results are retrieved from \cite{37}.}
\end{tabular}
}
\end{table}

\section{Discussions and Future Directions for RS Diffusion Models}
As discussed in the previous sections, diffusion models are rapidly evolving in the RS community, presenting great potential from generating RS images to enhancing the image quality, and further to recognition and detection. In fact, the research on diffusion models in RS is still at an early stage, with many tasks to be explored and further improvements to be achieved. In the following sections, we will discuss the possible future research directions from two aspects: extended applications and model training and deployment.

\begin{figure}[t]
\centering
\includegraphics[scale=0.68]{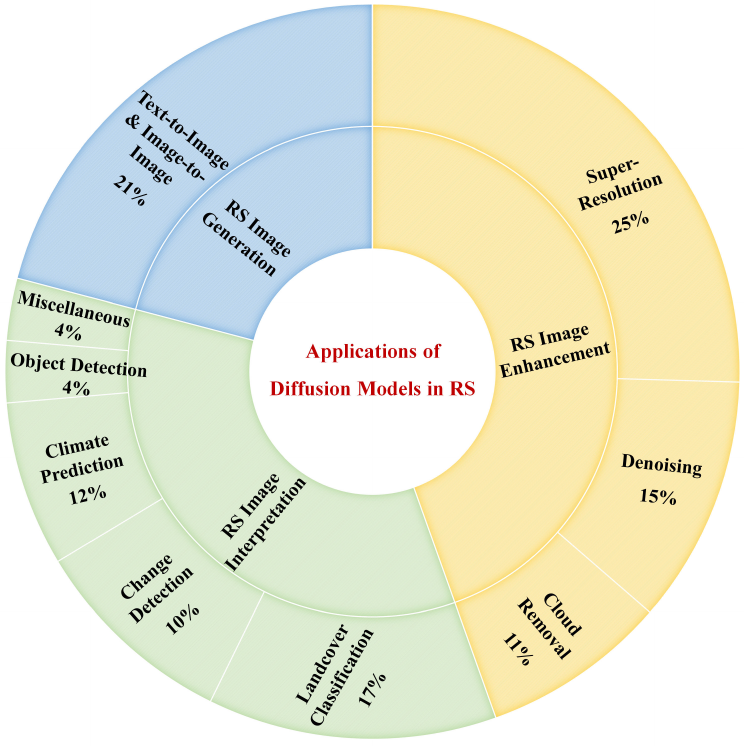}
\caption{Frequency of diffusion models in different RS applications.}
\label{pan_all}
\end{figure}

\subsection{Extended Applications}
\subsubsection{\textbf{For Specific RS Tasks}}
As shown in Fig. \ref{pan_all}, diffusion models are most frequently used in reconstructing high-resolution RS images (i.e., the super-resolution task), followed by the generation task, which accounts for 21\% of all reviewed papers. Apart from a few studies on object detection and miscellaneous tasks, the application of diffusion models in other remote sensing (RS) tasks remains relatively consistent. Fig. \ref{development} helps elucidate the reasons behind this phenomenon. Since 2024, research on the RS image generation task has significantly decreased. However, there has been a notable increase in research on RS image interpretation tasks, such as climate prediction \cite{101,102} and change detection \cite{88,89,87}. This shift suggests that researchers have recognized the limitations of the diffusion models in the RS community and have identified the development of more effective RS diffusion models for image interpretation tasks as an important research direction for the future.

While diffusion models are also commonly applied in landcover classification, there exists a significant limitation. In these methods, diffusion models mainly serve as feature extractors for the input image \cite{38,39,40}, requiring an additional classifier/detector to execute the specific task. Such separation of feature extraction and detection is not straightforward and usually needs two-step training, which is likely to fall into a suboptimal solution due to the extracted features not being fully suitable for the specific task. Therefore, future research could focus on developing RS task-specific diffusion models, by incorporating task-related prior in the model design or designing end-to-end models to improve the performance of diffusion models on specific RS tasks.

It is worth noting that the vast majority of existing RS diffusion models are based on the U-Net architecture, with a few works incorporating the Transformer architecture or its attention mechanism \cite{24,36,38,81,85,89}. Among them, only \cite{81} and \cite{36} employs a complete Transformer-based diffusion model, U-ViT \cite{U-ViT}, for the diffusion process. Nevertheless, the U-ViT consists of long skip residual connections, which is aligned with those of U-Net. In contrast, DiT \cite{DiT}, another diffusion model entirely based on the Transformer architecture, set the residual connections within each block, allowing the attention layer to perform global convolution and information extraction at a finer-grained scale, achieving state-of-the-art results in both image and video generation \cite{DiT,DiT-3D,DiT1, DiT2}. Thus, it can be seen that the diffusion model based on transformer architecture holds great potential for development, warranting further exploration by researchers in the field of RS.

\begin{table}[t]
\centering
\caption{Categorization of Diffusion Models in RS Based on Image Modalities}
\label{app_table} 
\renewcommand{\arraystretch}{1.3}
\setlength\tabcolsep{2.2mm}{
\begin{tabular}{ccc}
\hline
Modality                       & Application              & Correlation paper                                                                                                                                                                                                                                                                                             \\ \hline
\multirow{3}{*}{Multispectral}       & Generation               & \begin{tabular}[c]{@{}c@{}} \cite{7}\cite{2}\cite{70}\cite{3}\cite{12}\cite{811}\cite{4}\\ \cite{9}\cite{71}\cite{80}\cite{50}\cite{81}\cite{52}\cite{10}\end{tabular} \\ \cline{2-3} 
& Super-Resolution         &\begin{tabular}[c]{@{}c@{}} 
\cite{23}\cite{24}\cite{22}\cite{105}\cite{106}\cite{21}\\
\cite{25}\cite{108}\cite{111}\cite{112}\cite{113}                                                                                                                                                                                       \end{tabular}\\ \cline{2-3} 
             & Cloud Removal            &\begin{tabular}[c]{@{}c@{}}  \cite{44}\cite{54}\cite{42}\cite{43}\\
             \cite{94}\cite{93}\cite{41}\cite{95}                                                                                                        \end{tabular}\\ \hline
\multirow{3}{*}{Hyperspectral} & Super-Resolution         & \begin{tabular}[c]{@{}c@{}} \cite{34}\cite{32}\cite{107}\cite{30}\cite{fusion_new}\cite{26}\\ \cite{31}\cite{104}\cite{29}\cite{109}\cite{110}  \end{tabular}
                    \\ \cline{2-3} 
                               & Classification & \begin{tabular}[c]{@{}c@{}} \cite{38}\cite{85}\cite{40}\cite{39}\\ \cite{64}\cite{classHSI24}\cite{57}\cite{55}\cite{85} \end{tabular}        \\ \cline{2-3}  
                               & Denoising               & \cite{16} \cite{15}\cite{denoiseHSI24}                                                                                                                                                                                                                                                                         \\ \hline
\multirow{2}{*}{SAR}                             & Denoising                & \cite{18}\cite{17}\cite{78denoise_new0}\cite{19}\cite{79denoise_new1}                                                                                                                                                                                                                                                      \\ \cline{2-3} 
\multicolumn{1}{l}{}           & Generation               & \cite{13}\cite{51}\cite{11}                                                                                                                                                                                                                                                      \\ \hline
\end{tabular}
}
\end{table}

\subsubsection{\textbf{For Multi-Modal RS Images}}
Unlike natural images, RS images are captured by different types of sensors, encompassing multiple modalities. Table \ref{app_table} lists the top three most common applications across different modal RS images. 
Obviously, most of the diffusion model-based methods are developed for multispectral images. For HSI images, the applications of diffusion models are focused on super-resolution (especially pansharpening), and classification tasks. As for SAR images, diffusion models only appear in the process of generation and denoising. However, the spectral signatures of HSI and the robustness of SAR images play crucial roles in the recognition and detection tasks \cite{HSItarget0, HSItarget3,SARtarget3}. Therefore, exploring how to effectively apply diffusion models to multi-modal RS images is a necessary future research direction. In this way, the unique information of different modalities can be leveraged to improve the accuracy of RS image analysis.

In addition, LiDAR data, as an important type of RS data that can provide surface height information and ground structure details, is seldom used in the existing RS diffusion models. Only Li \emph{et al.} \cite{39} introduced LiDAR images as auxiliary information when using diffusion models for hyperspectral classification. In fact, LiDAR images can not only be used as a supplement to other modalities \cite{Lidar0}, but also generate continuous 3D terrain models for topographic and geomorphological analysis \cite{Lidar5}, vegetation detection \cite{Lidar2}, and urban planning \cite{Lidar3}. Recently, diffusion models have demonstrated satisfactory performance in 3D point cloud generation \cite{3dimage2,3dimage3}. Such technological advancements may be borrowed to the RS LiDAR data, thereby filling the gap of LiDAR data in various RS tasks.

\subsubsection{\textbf{For Realistic RS Images}}
Although many diffusion model-based RS image generation methods have been developed, they often overlook some special features of RS images, resulting in noticeable gaps between the synthesized image and real images. For example, diffusion model-based HSI generation methods usually need to compress the spectral dimension \cite{5,8}, which neglects the details of spectral curves and hinders the generation of accurate spectra. SAR images are complex in nature, comprising both amplitude and phase terms \cite{SARchallenge}. However, the generation methods developed so far have mainly focused on the image amplitude, ignoring the phase information. Therefore, future diffusion model-based generation methods should take these aspects into account to obtain more realistic RS images.

Another aspect that is often overlooked is the size of real RS images, which tend to be quite large (e.g., Gaofen-2 images are 29,200 $\times$ 27,620 pixels) \cite{largeRS1,largeRS2}. Existing diffusion model-based generation methods are primarily designed for patch-level images with 256 $\times$ 256 pixels, which means that a large-scale RS image can only be obtained by stitching multiple patch images \cite{10}. From the visual perspective, this approach is suboptimal since it is difficult to ensure the continuity of scenes and easy to leave traces at the joints. Thus, how to obtain realistic and reasonable large-scale RS images with diffusion models is a worthy research direction for further exploration. Notably, the increase in image size will inevitably bring computational and storage burdens. Accordingly, how to generate large-scale RS images under limited resources also needs careful consideration.

\subsubsection{\textbf{For General RS Model}}
Nowadays, more and more researchers are devoting themselves to developing general intelligent models, which can provide more accessible and high-performing solutions to help both industry professionals and interested non-professionals \cite{AGI1, AGI2, chatGPT, Dall-E, AGI4}. For example, ChatGPT \cite{chatGPT} has greatly simplified the process of collecting and summarizing information, while DALL-E \cite{Dall-E} has facilitated the rapid transformation of artistic ideas into practical examples. In the field of RS, some researchers are also attempting to develop universal multi-modal large models \cite{RSlarge1, RSlarge2, RSlarge3, RSlarge4}. However, these studies are still in the embryonic stage and have a lot of room for development. Fortunately, diffusion models have shown superior performance on various RS applications, as presented before. Therefore, a promising research direction is to construct a general RS intelligent model based on the diffusion model, which can span different modalities of RS images and accomplish multiple earth observation tasks.

\subsection{\textcolor{black}{Model Training and Deployment}}
It is widely acknowledged that diffusion models require a substantial number of iterations to generate high-quality samples, which is a noticeable drawback of this technology. \textcolor{black}{Moreover, these models often encompass a substantial number of parameters, necessitating training and deployment on devices with powerful neural computing units \cite{mobilediffusion, speed2023cvprw}.} However, the resources on satellites are extremely limited and lack the computational power for such requirements. What's more serious is that the deep-space environment is severe, often subject to extreme climate and illumination changes, which puts higher demands on the stability and reliability of the model. \textcolor{black}{Therefore, exploring more efficient ways to train RS diffusion models and deploying them on resource-limited satellites within harsh environments are highly meaningful research directions.}

\subsubsection{\textbf{Accelerate Processing}}
\textcolor{black}{Given the huge computational cost of training diffusion models from scratch, a common training paradigm is to fine-tune pre-trained diffusion models. Traditional fine-tuning methods involve updating all the parameters of the pre-trained model, which carries a high risk of overfitting as the size of pre-trained models increases. To address this problem, researchers have proposed a series of fine-tuning methods, such as inserting network modules with a small number of learnable parameters (i.e., adapters) \cite{Lora, adaptformer, difffit}, or selecting only a part of the model parameters to update \cite{bitfit, scaling, svdiff}. 
One of the most popular methods is Low-Rank Adaptation (LoRA) \cite{Lora}, which is also applied to train diffusion models for the RS image generation task. However, the training performance of LoRA still lags significantly behind full parameters fine-tuning, and other fine-tuning methods have not been applied to the RS diffusion models. Therefore, how to apply pre-trained diffusion models in the RS domain with minimal parameter updates is still worth exploring.}

As for the inference process of diffusion models, there are many accelerated sampling approaches that have been proposed \cite{speed0, speed1, DDIM, DPM-solver, speed2}. These methods have successfully reduce the necessary sampling steps from several hundred to dozens, or even a few or a single step \cite{onestep1, onestep2, onestep3}. However, Tuel \emph{et al.} \cite{11} found that these accelerated sampling methods, such as DDIM \cite{DDIM} or DPM-solver \cite{DPM-solver}, did not perform well on SAR images. This observation suggests that the significant differences between RS images and natural images make these acceleration methods, originally designed for natural images, inapplicable to RS imagery. More recently, Kodaira \emph{et al.} \cite{streamdiffusion} proposed StreamDiffusion, which achieves the speed of generating images up to 91.07fps on a 4090GPU through pipelined batch processing. Unfortunately, they did not evaluate the proposed model on RS images and overlooked that some of the acceleration techniques are not suitable for devices with limited computing resources. It can be seen that developing acceleration methods for RS diffusion models is still an open problem.

Moreover, designing lightweight diffusion models is also an effective way to reduce training and inference costs, which can be specifically achieved by distillation techniques \cite{4, 24, distillation1, distillation2} or changing the network structure \cite{21, 22, light1, light2}. Nevertheless, the more prevalent solution in RS is to place the diffusion model in a lower-dimensional latent space through data compression, like \cite{31,29,8,67}, which is inspired by the LDM, thereby shrinking the sampling space and reducing computational overhead. However, the performance of such methods is limited by the quality of the constructed latent space \cite{DM11}. If the latent space is incapable of extracting useful semantic information from RS imagery, or critical information is lost during the dimensionality reduction process, the final generated RS images will be adversely affected. Therefore, it is necessary to explore better lightweight structure design methods than compressing the data space.

\subsubsection{\textbf{Improve Stability and Reliability}}
In order to keep the diffusion model run stably in the harsh environment, Yu \emph{et al.} \cite{20} proposed a diffusion model-based adversarial defense approach to protect deep neural networks from a variety of unknown adversarial attacks, effectively improving the robustness of the model.
Similar research on employing diffusion models to tackle the complex environments in RS should be further explored, especially to design specific diffusion models for different environmental disturbances or different sensors. Such efforts would contribute to the deployment of diffusion models in various environments and ensure their long-term operation, holding great practical value.

In addition, with the rapid development of Global Observation System (GOS), the distributed learning of intelligent models over multiple satellites has gradually become one of mainstream directions in RS \cite{distributed0, distributed01, distributed1, distributed2}. More recently, Li \emph{et al.} \cite{distrifusion} reported that the speed of generating images with diffusion models can be effectively improved under this parallel learning mode, further confirming prospects of distributed diffusion models in RS. However, when deploying diffusion models in such multi-client distributed scenarios, ensuring the security of RS data is an essential issue \cite{security}. Consequently, developing robust and trustworthy diffusion models has become an urgent necessity in the RS community.

\section{Conclusion}
In summary, the emergence of diffusion models has created a new era of intelligent RS image processing. Compared to other deep generative models, diffusion models are robust to the inherent noise in RS images, better adapt to their variability and complexity, and offer a more stable training process. Hence, applying diffusion models to various RS tasks has become an inevitable trend.
For this reason, this paper first introduces the theoretical background of diffusion models to help understand how the diffusion model works for RS tasks. Then, it reviews and summarizes studies on the use of diffusion models in processing RS images, including image generation, super-resolution, cloud removal, denoising, and a series of interpretation tasks such as landcover classification, change detection, and climate prediction. Moreover, the paper takes cloud removal, landcover classification, and change detection as examples to demonstrate the superiority of diffusion models in various RS image processing tasks through visual results and quantitative indicators. Finally, the paper discusses the limitations of the existing diffusion models in RS and highlights that further exploration could be carried out on the extended applications and model deployment. We hope this paper can provide a valuable reference for researchers in related fields to stimulate more innovative studies to break the performance bottleneck of existing methods or to promote the development of diffusion models for more RS applications.

\section*{\textcolor{black}{Acknowledgment}}
\textcolor{black}{The authors sincerely appreciate the valuable feedback and constructive suggestions provided by the editor-in-chief, associate editor, and reviewers, which have notably enhanced the quality of this paper.}

\textbf{}
\ifCLASSOPTIONcaptionsoff
  \newpage
\fi

\bibliographystyle{IEEEtran}
\bibliography{IEEEabrv,RSdiff}

\begin{IEEEbiography}[{\includegraphics[width=1in,height=1.25in,clip,keepaspectratio]{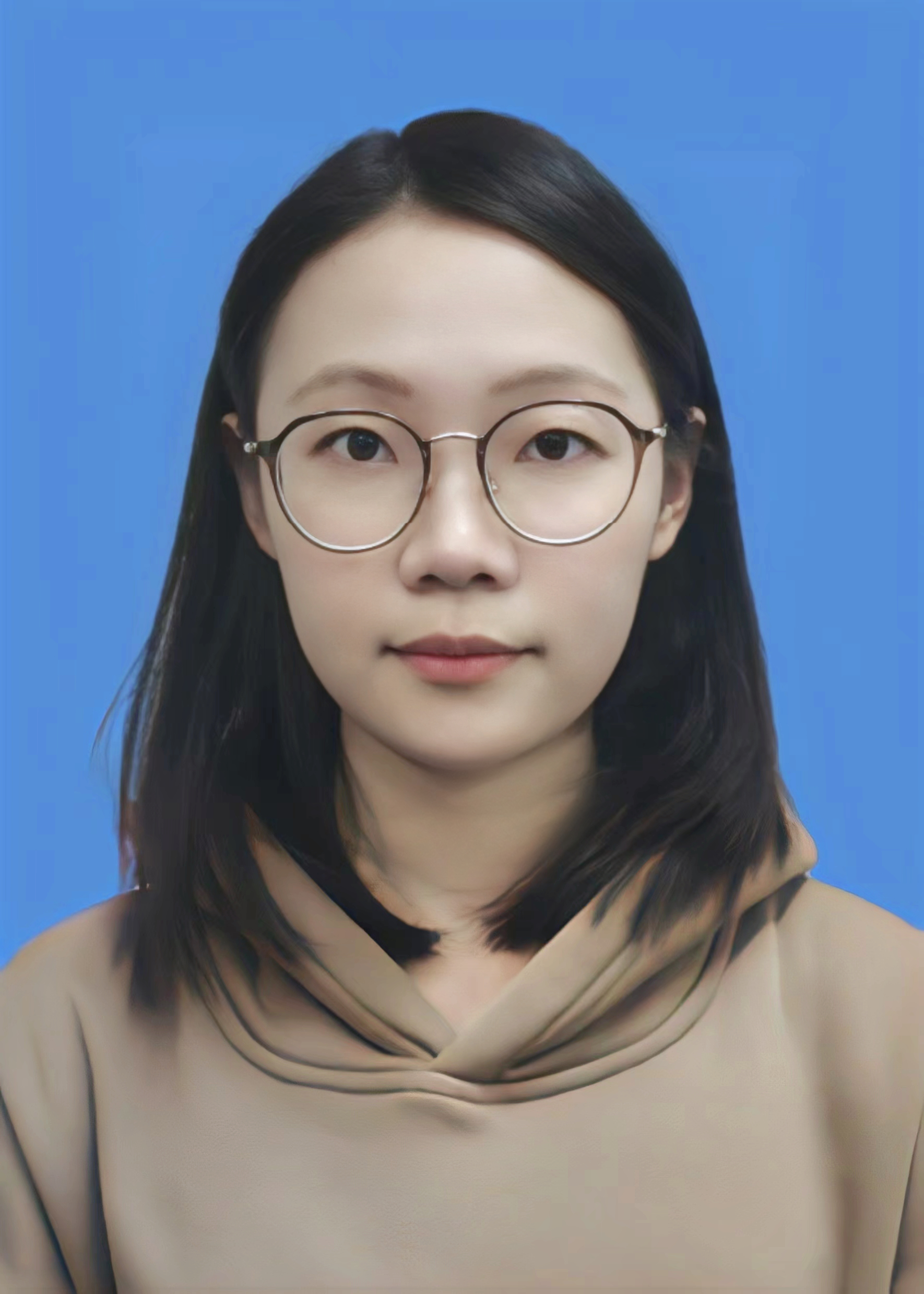}}]{Yidan Liu} received the B.E. degree and M.S. degree in telecommunications engineering from the Xidian University, Xi’an, China, in 2020 and 2023. She is currently pursuing her Ph.D. degree at the Laboratory of Vision and Image Processing, Hunan University, Changsha, China. \par
Her research interests focus on deep learning, image processing and computational imaging.
\end{IEEEbiography}

\begin{IEEEbiography}[{\includegraphics[width=1in,height=1.25in,clip,keepaspectratio]{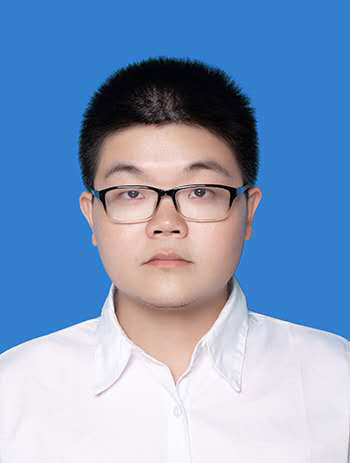}}]{Jun Yue} received his B.Eng. degree in geodesy from Wuhan University, Wuhan, China, in 2013 and his Ph.D. degree in geographic information systems from Peking University, Beijing, China, in 2018. He is currently an assistant professor at the School of Automation, Central South University, Changsha 410083, China. \par
His research interests include satellite image understanding, pattern recognition, and few-shot learning. He is a reviewer for {\scshape IEEE Transactions on Image Processing}, {\scshape IEEE Transactions on Neural Networks and Learning Systems}, {\scshape IEEE Transactions on Geoscience and Remote Sensing}, ISPRS Journal of Photogrammetry and Remote Sensing, {\scshape IEEE Geoscience and Remote Sensing Letters}, {\scshape IEEE Transactions on Biomedical Engineering}, \textit{Information Fusion}, and \textit{Information Sciences}.
\end{IEEEbiography}

\begin{IEEEbiography}[{\includegraphics[width=1in,height=1.25in,clip,keepaspectratio]{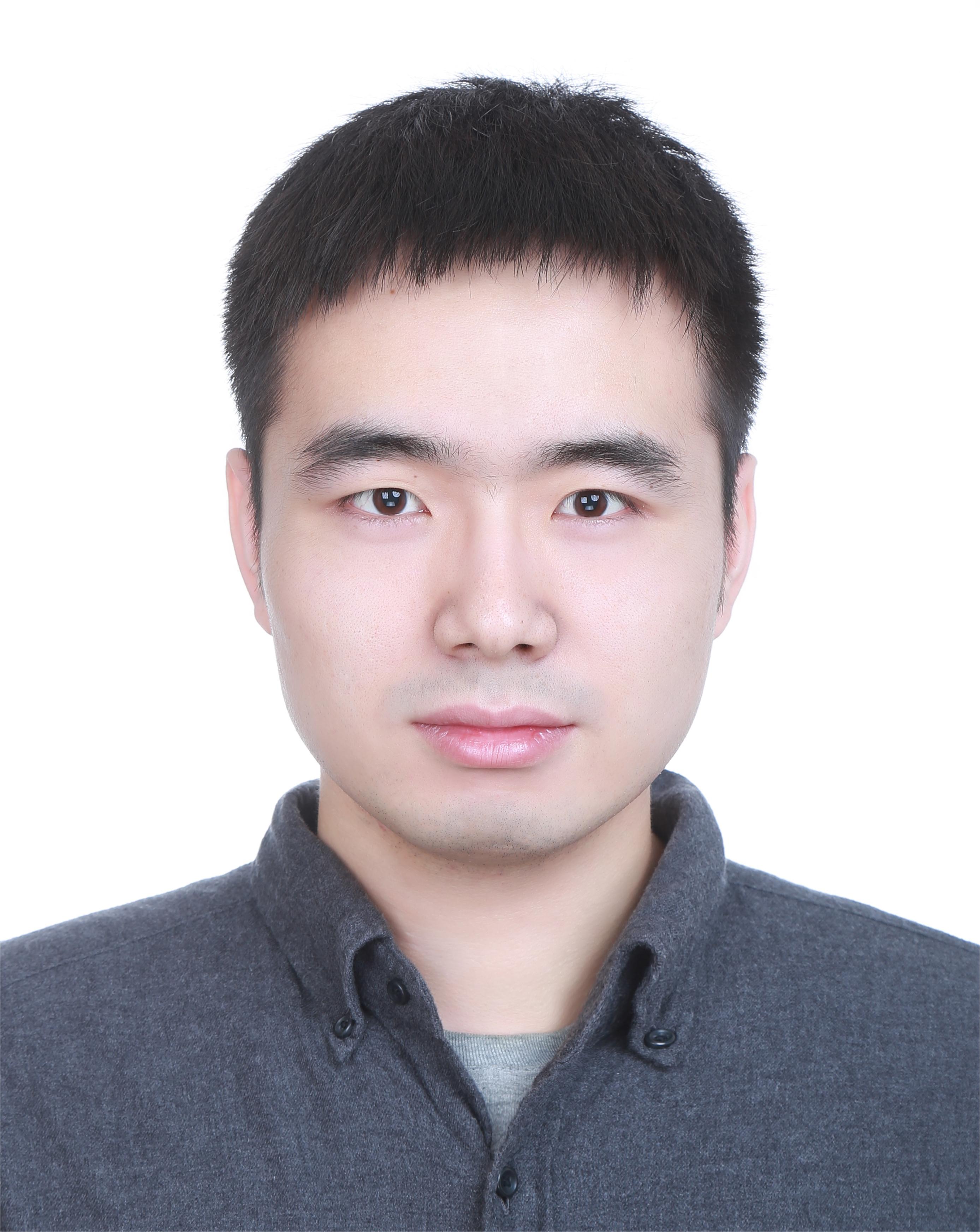}}]{Shaobo Xia} received the bachelor's degree in geodesy and geomatics from the School of Geodesy and Geomatics, Wuhan University, Wuhan, China, in 2013, the master's degree in cartography and geographic information systems from the University of Chinese Academy of Sciences, Beijing, China, in 2016, and the Ph.D. degree in geomatics from the University of Calgary, Calgary, AB, Canada, in 2020. \par
He is an Assistant Professor with the Department of Geomatics Engineering, Changsha University of Science and Technology, Changsha, China. His research interests include Geomatics and 3D Deep Learning.\end{IEEEbiography}

\begin{IEEEbiography}[{\includegraphics[width=1in,height=1.25in,clip,keepaspectratio]{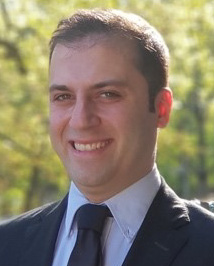}}]{Pedram Ghamisi} received his Ph.D. degree in electrical and computer engineering at the University of Iceland in 2015. He is currently the head of the machine learning group at Helmholtz-Zentrum Dresden-Rossendorf, Helmholtz Institute Freiberg for Resource Technology, Freiberg, 09599 Germany, and a visiting professor and the group leader of AI4RS at the Institute of Advanced Research in Artificial Intelligence, Vienna, 1030, Austria. He is a cofounder of VasoGnosis with two branches in San Jose, CA, USA, and Milwaukee, WI, USA. He was cochair of the IEEE Image Analysis and Data Fusion Committee (IADF) between 2019 and 2021. He was a recipient of the IEEE Mikio Takagi Prize for winning the Student Paper Competition at the IEEE International Geoscience and Remote Sensing Symposium in 2013, the first prize of the data fusion contest organized by the IADF in 2017, the Best Reviewer Prize of IEEE Geoscience and Remote Sensing Letters in 2017, and the IEEE Geoscience and Remote Sensing Society 2020 Highest Impact Paper Award. His research interests include interdisciplinary research on machine (deep) learning, image and signal processing, and multisensor data fusion. He is a Senior Member of IEEE. For more information, please see http://pedram-ghamisi.com/.
\end{IEEEbiography}

\begin{IEEEbiography}[{\includegraphics[width=1in,height=1.25in,clip,keepaspectratio]{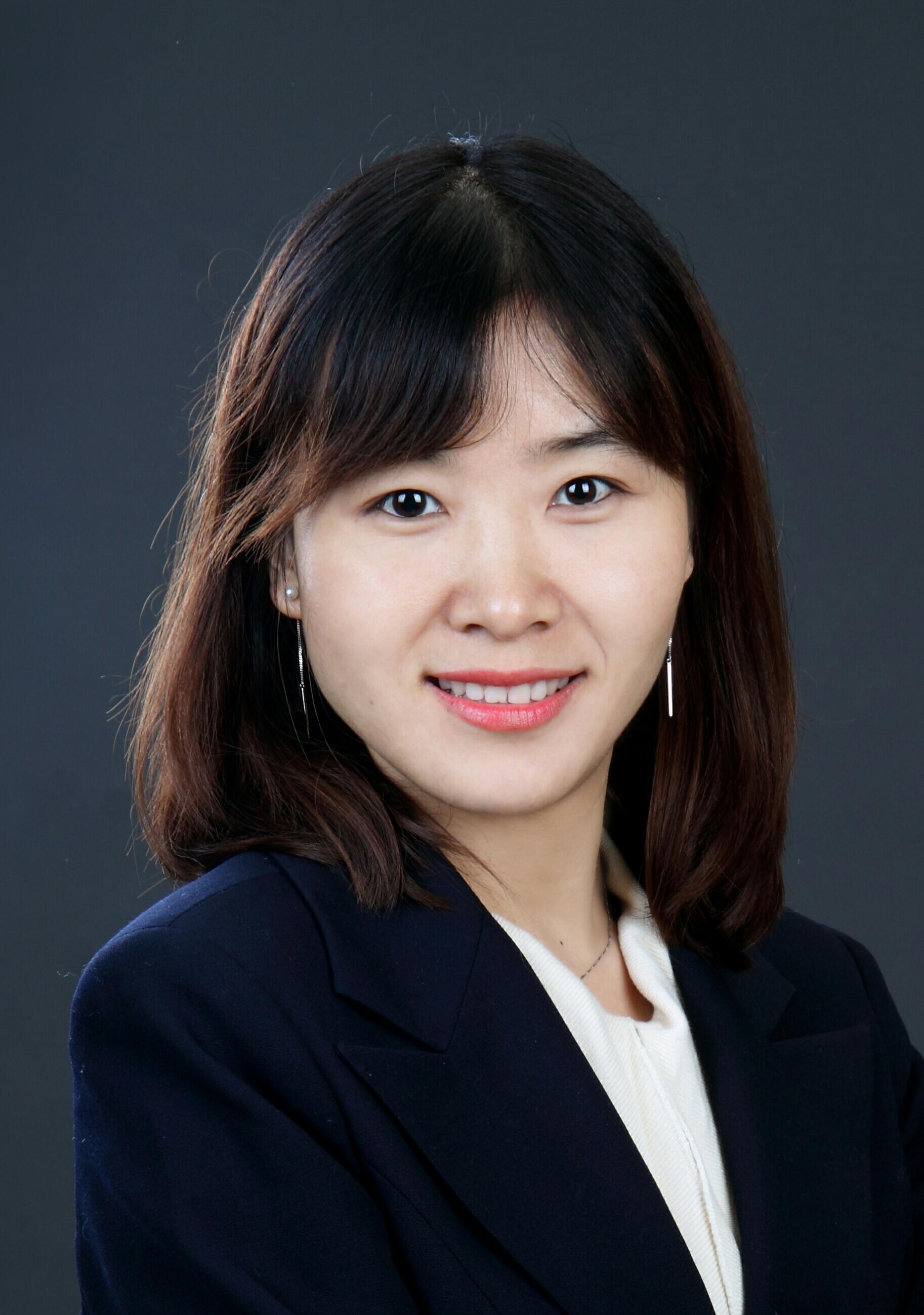}}]{Weiying Xie} (Senior Member, IEEE) received the Ph.D. degree in communication and information systems from Xidian University, Xi’an, China, in 2017.
Currently, she is a Professor with the State Key Laboratory of Integrated Services Networks, Xidian University. She has published more than 50 articles in refereed journals and proceedings, including {\scshape IEEE Transaction on Image Processing}, {\scshape IEEE Transactions on Geoscience and Remote Sensing}, {\scshape IEEE Transactions on Neural Networks and Learning Systems}, {\scshape IEEE Transaction on Cybernetics}, and Conference on CVPR and AAAI. Her research interests include neural networks, machine learning, hyperspectral image processing, and high-performance computing.
\end{IEEEbiography}

\begin{IEEEbiography}[{\includegraphics[width=1in,height=1.25in,clip,keepaspectratio]{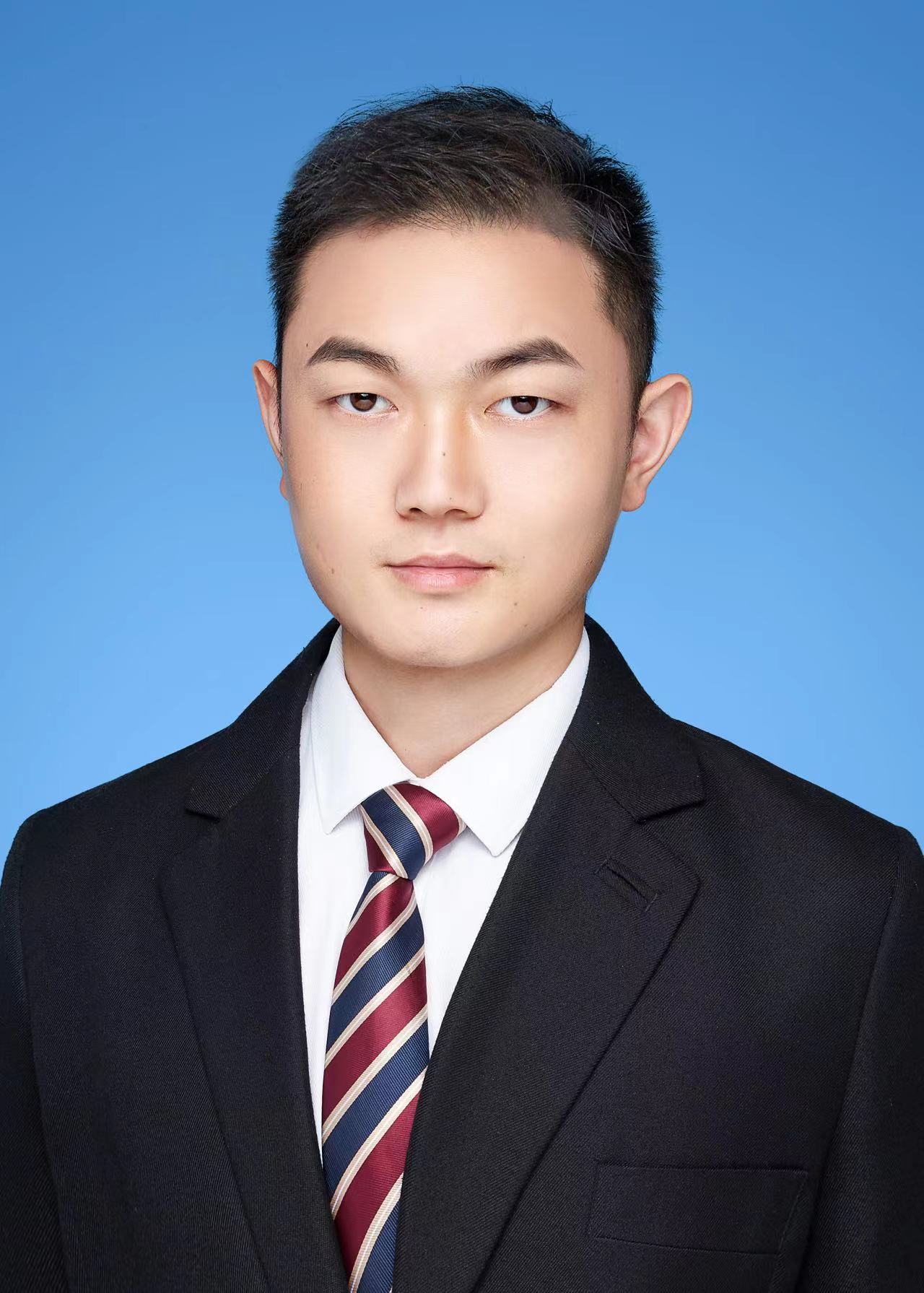}}]{Leyuan Fang} (Senior Member, IEEE) received the Ph.D. degree from the College of Electrical and Information Engineering, Hunan University, Changsha, China, in 2015.
	
From September 2011 to September 2012, he was a Visiting Ph.D. Student with the Department of Ophthalmology, Duke University, Durham, NC, USA, supported by the China Scholarship Council. From August 2016 to September 2017, he was a Post-Doctoral Researcher with the Department of Biomedical Engineering, Duke University. He is a Professor with the College of Electrical and Information Engineering, Hunan University. His research interests include sparse representation and multiresolution analysis in remote sensing and medical image processing. 

Dr. Fang was a recipient of the 2nd-Grade National Award at the Nature and Science Progress of China in 2019. He is an Associate Editor of {\scshape IEEE Transactions on Image Processing}, {\scshape IEEE Transactions on Geoscience and Remote Sensing}, {\scshape IEEE Transactions on Neural Networks and Learning Systems}, and \textit{Neurocomputing}.
\end{IEEEbiography}

\end{document}